\documentclass[10pt,twocolumn,letterpaper]{article}
\usepackage{iftex}
\ifXeTeX\usepackage[T1]{fontenc}\fi %
\usepackage[pagenumbers]{cvpr}      %
\usepackage{multirow}
\usepackage{tikz}
\usetikzlibrary{patterns,arrows.meta,positioning,calc}

\usepackage{enumitem}\usepackage{placeins}\usepackage{algorithm}\usepackage{algpseudocode}

\definecolor{cvprblue}{rgb}{0.21,0.49,0.74}
\usepackage[breaklinks,colorlinks,allcolors=cvprblue]{hyperref}
\PassOptionsToPackage{nameinlink}{cleveref}
\definecolor{rootblue}{RGB}{0,114,178}               %
\newcommand{\gain}[1]{\textsubscript{\textcolor{rootblue}{#1}}}   %
\newcommand{\dropd}[1]{\textsubscript{\textcolor{gray}{#1}}}       %
\newcommand{\up}{\,$\uparrow$}
\newcommand{\safe}{\textcolor{rootblue}{\checkmark}}
\newcommand{\unsafe}{\textcolor{BrickRed}{$\times$}}
\newcommand{\pp}{\,pp}
\newcommand{\vfig}[1]{\includegraphics[width=0.95\textwidth]{fig/#1.pdf}}

\title{Label-free Self-repair of Visual Fragility in World Action Models}
\author{%
Zhiruo Zhou$^{1,2}$\quad Xiwen Chen$^{1,3}$\quad Zelin Li$^{1}$\quad Yinghao Wang$^{4}$\quad Xiaojun Zhu$^{1}$\quad Houde Liu$^{1}$\\[2pt]
$^{1}$Tsinghua University\quad $^{2}$Wuhan University of Technology\\
$^{3}$Shanghai Jiao Tong University\quad $^{4}$Harbin Institute of Technology}

\begin{document}
\twocolumn[{%
\renewcommand\twocolumn[1][]{#1}%
\maketitle
\vspace{-7mm}
\begin{center}
\captionsetup{type=figure}
\vfig{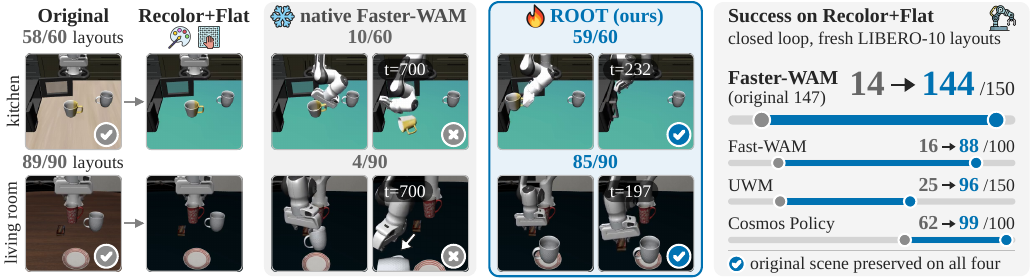}
\vspace{-1mm}
\captionof{figure}{\textbf{Once the working surface changes, world action models fail; ROOT automatically restores them.} Left: on two fresh LIBERO-10 layouts, native Faster-WAM succeeds in the original scene but fails when only the work surface is recolored and flattened; ROOT completes both layouts. Right: closed-loop success on Recolor+Flat, with original-scene success preserved.}
\label{fig:teaser}
\end{center}%
\vspace{-2.5mm}
}]

\begin{abstract}
When a world action model (WAM) drops from 147 successes to 14? It is to change only the color and texture
of the table. A near-perfect WAM is paralyzed. Significantly, we find that, in most WAMs, the site where such a visual shift is ``read out'' tends to lie in the later half of the network. Accurately inserting a lightweight adapter at the input stage achieves a repair effect close to that of full-network fine-tuning. Building on this, we propose ROOT (\underline{R}epair at the \underline{O}rigin with \underline{O}wn \underline{T}argets). ROOT forces the model to shift to its original predictions on visual fragility conditions, enabling normal task completion across diverse appearances. Remarkably, ROOT is label-free without action labels, target data, or architecture modifications. Repair location, training duration, and deployment strength are all automatically configured by ROOT. ROOT lifts Faster-WAM's success on recolored, flattened tables from single digits to near-perfect. On LIBERO-Plus, ROOT achieves the largest gain to date for a post-hoc repair (+25.9pp). For Streaming-WAM, under unseen conditions, and on GE-Act on CALVIN in a second simulator, ROOT yields gains of 55.8\,pp and 21.6\,pp. On real robots, the success rate is raised from 36.7\% to 93.3\% over 180 trials.
\end{abstract}

\begin{figure*}[t]
\centering
\includegraphics[width=\textwidth]{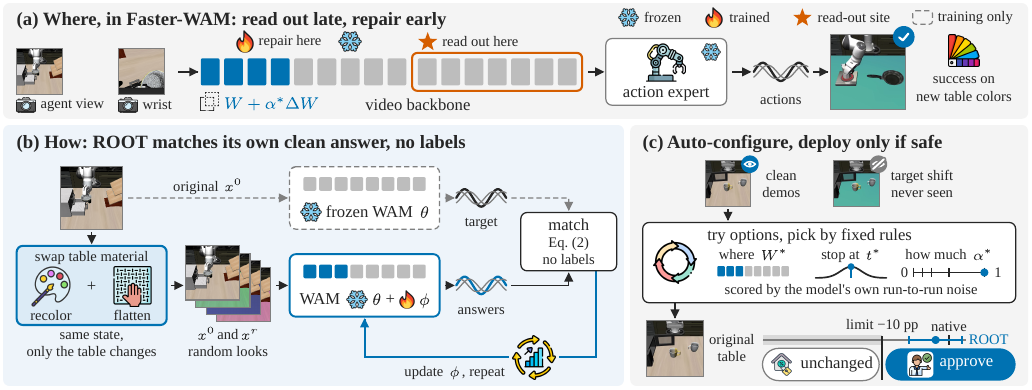}
\vspace{-5mm}
\caption{\textbf{ROOT on Faster-WAM: read the shift out late, repair early, deploy only if the original scene holds.} (a)~Restoration locates the read-out in late blocks ($\star$, \cref{sec:locate}); ROOT adapts early blocks. (b)~The frozen model's clean-view prediction is the target for $x^0$ and for random looks $x^r$ (recolored, flattened) under shared noised actions $a_\tau$; dashed: training only. (c)~From clean data alone, fixed rules pick the window, checkpoint and scale, and deploy only with a non-inferior original scene.}
\label{fig:method}
\vspace{-3mm}
\end{figure*}

\section{Introduction}
\label{sec:intro}

World action models (WAMs) generate future video and actions asynchronously, or let an action expert read the internal state of a
pre-trained video model \cite{fasterwam,fastwam,cosmospolicy,uwm}, and approach ceiling success on LIBERO
\cite{libero}. A deployed robot, however, meets a new background, light or camera far more often than a new task, and changing
only the work surface's visual appearance can break a WAM \cite{wamrobust,liberoplus,stwam}: recoloring and flattening the table
turns a policy that succeeds in 147 of 150 layouts into one that succeeds in 14, with task and physical state unchanged
(\cref{fig:teaser}). Robustness suites \cite{liberoplus,liberopro} measure such losses, and existing remedies retrain with
new branches or external semantic targets \cite{stwam,robustwam}. Whoever holds a released WAM, however, has the model and
its clean data, not a new architecture or data from the shifted scene. \emph{Can a WAM repair its own visual fragility (appearance
brittleness) from what it already has?}

Where a visual appearance shift is read out is not effectively and efficiently where it should be repaired. We build SurfaceShift, a color-calibrated work-surface protocol of paired renderings and fresh layouts that separates color,
lightness and texture (\cref{sec:setup,sec:diagnosis}), and with it privileged restoration, which substitutes activations from the same state in the original scene, shows where each
model reads the shift out: late video values in Faster-WAM, late keys and values in Fast-WAM, the wrist-image input in
Cosmos Policy (\cref{fig:readout}). Repair probes disagree with these read-out maps in a consistent way: on all three
models, adapters on the first blocks recover most of a full adapter's effect, late-only adapters far less
(\cref{sec:locate}). The localization/editing gap of offline language models \cite{hase2023localization} thus
reappears in closed-loop control and shows where to act.

In this paper, ROOT (Repair at the Origin with Own Targets) makes the existing WAM's prediction for the clean view the
target for every perturbed view, so it needs no action labels, no data from the target shift and no architectural change;
an automatic \emph{locate-then-repair} procedure selects the adapter window, checkpoint and deployment scale with fixed
thresholds and deploys only if the original scene is preserved (\cref{fig:method}). Stability training regularizes a model
while it is trained \cite{zheng2016stability}; but ROOT repairs the model to complete robotic tasks with higher success rates.

\paragraph{Contributions.}
\begin{itemize}[leftmargin=*,itemsep=1pt,topsep=2pt]
\item \textbf{We locate where to repair a visual shift.} SurfaceShift separates color, lightness and texture; a WAM reads a shift out
  late, yet adapters on the first blocks recover most of a full adapter's effect (\cref{fig:factors,fig:readout}).
\item \textbf{Against this collapse, we propose ROOT.} Its target is the WAM's own clean-view prediction for every perturbed view: no
  labels are needed, the original scene has nothing to change, whereas demonstration actions pull the policy toward the demonstrator.
  Deployment is fully automatic (\cref{sec:method}).
\item \textbf{ROOT generalizes further than same-budget repairs.} Recolor+Flat success on Faster-WAM rises from 9.3 to 96.0\%, above
  privileged restoration. On unseen textures it beats every same-budget repair by 12 to 21 points, and carries over to a model unused
  in development, a second simulator and a real robot (36.7 to 93.3\%)
  (\cref{sec:experiments,sec:boundaries}).
\end{itemize}

\section{Related Work}
\label{sec:related}

\paragraph{WAMs and their visual fragility.} Video-generation-based policies connect
predicted visual dynamics to control
\cite{videopolicy,genieenvisioner,dreamzero}. Robustness
suites \cite{liberoplus,liberopro,colosseum,factorworld} and a WAM study
\cite{wamrobust} document their visual fragility. WAM-specific remedies introduce
semantic branches \cite{stwam}, align action or video representations
with semantic features \cite{robustwam,agra}, or enforce consistency
across camera views \cite{scvc}. None of them repairs from what the model already has. Our repair keeps the architecture and uses the model's own prediction as its only target.

\paragraph{Localizing versus repairing.} Activation patching and causal
tracing restore activations from a clean run to find the components that
carry a behavior \cite{vig2020causal,meng2022rome,zhang2024patching},
also in diffusion models \cite{basu2024localizing}; related analyses locate
vital layers, heads, or steering directions in diffusion transformers and robot policies
\cite{avrahami2025stableflow,haon2025vlasteer,mitra2025mechft}. Hase \etal\ \cite{hase2023localization} showed with offline
metrics that the localized layer does not predict where a factual edit
works best; surgical fine-tuning favors early layers for input-level
corruptions \cite{lee2023surgical} and last-layer retraining suffices for
spurious features \cite{kirichenko2023lastlayer}. We measure the
dissociation for appearance repair in WAMs by closed-loop
success and turn it into a placement rule that never reads the target shift. Steering of WAMs \cite{wamsteer} adds contrastive activation
directions at inference time; we train a small adapter once, at the input side (compared in \cref{sec:baselines}).

\paragraph{Consistency and robust fine-tuning.} Our objective builds on
stability training and AugMix \cite{zheng2016stability,hendrycks2020augmix}; consistency and augmentation improve
classifiers, visual RL agents and vision-language models \cite{goel2021modelpatching,svea,raileanu2021drac,nopd}. RobustVLA trains on perturbed inputs with demo-action targets \cite{robustvla}, anchoring keeps features near the base model \cite{anchoralign}, test-time methods adapt to the shift itself
\cite{tent,zhang2022memo,hansen2021pad}, and BYOVLA intervenes on observations without fine-tuning
\cite{hancock2024byovla}. Color or scene augmentation
\cite{tobin2017domain,rad,teoh2024greenaug,chen2023genaug} can erase color information
\cite{chromaguard}. Our scale interpolates weights as WiSE-FT, PAINT and RETAIN do \cite{wortsman2022wiseft,ilharco2022paint,retain}. ROOT instead takes the frozen
model's own clean-view prediction as the target for every perturbed view, so it needs no action labels and no data from the
target shift, and keeps the update only if the original scene is preserved. Concurrent PICA-WAM \cite{picawam} aligns perturbed-view
features with an EMA teacher during WAM training; ROOT leaves training untouched and repairs a released model post hoc, with an input-side
window chosen after the read-out diagnosis and a controller that deploys only if the original scene holds.

\paragraph{Where to adapt.} Adapters usually occupy every block \cite{houlsby2019adapters,lora}; placement is a design
dimension \cite{he2022unified,zhang2023adalora,xie2023difffit,zuo2026lorapatching}, and full fine-tuning can distort features
\cite{kumar2022lpft}. We test placement proxies against trivial baselines, following zero-cost
architecture search \cite{abdelfattah2021zerocost}. Concurrent work reports WAM brittleness
\cite{liberovpro,cswam}, edits activations \cite{yao2026rolloutfaith} or adapts at test time
\cite{roboirs,salt}. Closest, Not All Layers Need
Tuning \cite{notalllayers} allocates LoRA rank across VLA modules from training-free signals and fine-tunes on the target
environment; we place windows inside one WAM with a label-free objective that never sees the shift, judged in closed loop.

\section{Setup}
\label{sec:setup}

\paragraph{Models.} Seven WAMs in three backbone families.
\emph{Wan-based}: Faster-WAM \cite{fasterwam}, Fast-WAM \cite{fastwam}
and LingBot-VA \cite{lingbotva} share the Wan2.2-5B video DiT of the Wan family \cite{wan}
and differ in the action interface; Light-WAM \cite{lightwam} decodes actions from adapted features of the smaller Wan2.1-1.3B
and builds on Fast-WAM's code. \emph{Cosmos-Predict-based}: Cosmos
Policy \cite{cosmospolicy} decodes actions as latent slots of the video
sampler; DiT4DiT \cite{dit4dit} feeds features of the 18th block of the official
Cosmos-Predict2.5-2B DiT \cite{cosmospredict25} to a separate action DiT.
\emph{Encoder-based}: UWM \cite{uwm} sees images through a ResNet-18.
We use released LIBERO checkpoints unchanged (\emph{native}); for UWM, which has none for
LIBERO-10, we fine-tune it with the official recipe. Faster-WAM is the primary model; DiT4DiT and Light-WAM are the blind tests: neither shaped the procedure applied to it.
The frozen procedure alone configures two more: Streaming-WAM (LIBERO) and GE-Act (CALVIN).

\paragraph{The SurfaceShift protocol.} \emph{SurfaceShift} packages our
conditions, layouts and metrics: five mechanism-defined conditions; 112
appearance cells calibrated to target CIE $L^*a^*b^*$ values, so that
lightness, hue, chroma and texture can be read separately; paired fresh
layouts with fixed seeds; paired tests, original-scene non-inferiority
and the noise-envelope metric Env (\cref{sec:method}); and native results
for eight policies (datasheet and code: Supp.~S2; construction notes for the figures: Supp.~S8).

\paragraph{Conditions.} Every condition re-renders a LIBERO-10 task at a
fixed physical state and changes only the work-surface material. \textbf{Flat} replaces the texture by its mean color;
\textbf{Recolor} multiplies the material by a fixed blue color and keeps the texture; \textbf{Recolor+Flat}
does both. Under each scene's lighting, Recolor+Flat gives a bright teal surface in the four kitchen tasks (a hue shift at similar
lightness) and a near-black one in the six living-room and study tasks (a collapse of lightness; Supp.~S1.1). \textbf{Original} is the
unmodified scene; \textbf{Red-object} recolors a task-specific object red and says so in the instruction; both are safety
conditions.

\paragraph{Closed-loop protocol.} We fix a model-specific inference contract: horizon, re-planning interval and
the official fixed inference noise (Supp.~S1). The 50 official initial states per task were used during development, so we evaluate on
\emph{fresh layouts} from LIBERO's own sampler, generated without querying any model: 15 per task, 150 pairs (100 for Fast-WAM
and Cosmos Policy). All arms of a pair share one process, state and inference settings, and every arm uses the same layouts and seeds.
The layout counts in \cref{fig:teaser} cover all layouts of each appearance group.

\paragraph{Statistics.} The unit is the (task,
layout) pair. Primary tests are one-sided exact sign tests on discordant
pairs \cite{mcnemar1947} with Holm correction within each pre-registered
family \cite{holm1979}; p-values are Holm-adjusted within each family (families and raw values: Supp.~S5). Safety is non-inferiority at $-10$\pp\ (one-sided 95\% lower bound; Supp.~S1.2). Every closed-loop test has its
protocol, checkpoint and analysis frozen before the run \cite{nosek2018prereg}; $^\star$ marks privileged references that
read the same state in the original scene and are diagnostic tools.

\section{Diagnosis: Which Factors, and Where}
\label{sec:diagnosis}

\begin{figure*}[t]
\centering
\vfig{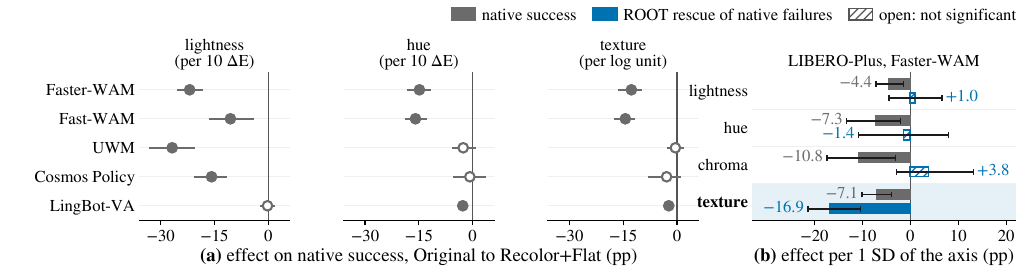}
\vspace{-1mm}
\caption{\textbf{Different WAMs fail along different appearance axes.} (a) Per-model effect of each axis on native success across SurfaceShift. (b) Faster-WAM, LIBERO-Plus Background/Light: per-SD effect on native success and ROOT's rescue; only texture predicts rescue.}
\label{fig:factors}
\vspace{-3mm}
\end{figure*}

We first ask which work-surface property breaks which model (\cref{fig:factors}), then where in the network it is read out.

\subsection{Models break along different axes}
\label{sec:factors}

Recolor+Flat combines several appearance changes, so its aggregate loss alone cannot identify what breaks a model.
\Cref{fig:factors}a separates these effects: lightness tracks the losses of Cosmos Policy and UWM, whereas Faster- and
Fast-WAM respond to lightness, hue and texture. DiT4DiT's preserved performance under Flat but collapse under recoloring
provides a further contrast; LingBot-VA changes comparatively little. A pre-registered sweep that varies lightness, hue
and texture on the same tasks confirms the lightness explanation: darkening the kitchen surface drives Cosmos Policy and
UWM from 38 and 35 successes to 0 of 40, and nearly all offline action tests respond more to lightness than to hue
(Supp.~S3). A repair must therefore cover more than color.

\subsection{Chroma correction fixes color, not texture}
\label{sec:chroma}

Following color constancy \cite{buchsbaum1980}, we construct a zero-training \emph{chroma correction}: it moves the
image's mean $a^*b^*$ back to that of the demonstrations while preserving lightness and texture. This intervention
isolates the color component. On Faster-WAM it recovers over half of the Recolor action deviation but essentially none of Flat (Supp.~S3.1), and
the closed loop agrees (\cref{tab:alt}): image statistics fix color, remaining failures track texture, so the
model must change.

\subsection{Where the shift is read out}
\label{sec:locate}

\begin{figure*}[t]
\centering
\vfig{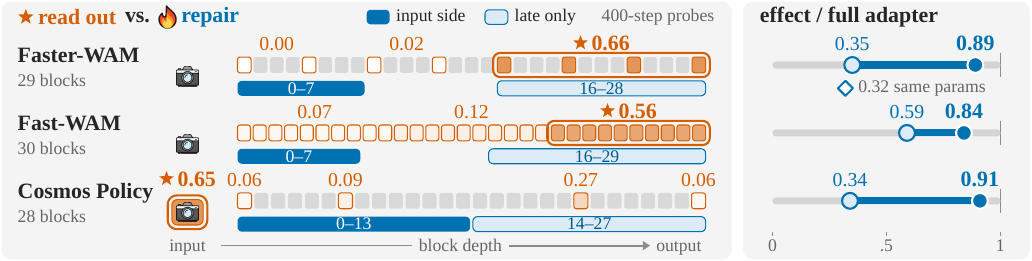}
\vspace{-1mm}
\caption{\textbf{Read out here, repair there.} Left: where privileged same-state restoration$^{\star}$ recovers the clean action (orange, $\star$ strongest site); the two ROOT windows tested (blue). Right: what each window recovers relative to the full adapter in 400-step probes. The shift is read out late in Faster- and Fast-WAM and at the image input in Cosmos Policy; input-side windows win on all three.}
\label{fig:readout}
\vspace{-3mm}
\end{figure*}

To localize the shift we run the perturbed episode but substitute selected
video-cache tensors from the same state rendered in the original scene at
each query. \Cref{fig:readout} shows read-out sites (Supp.~S3.3):
late video values in Faster-WAM, late keys and values in Fast-WAM, and the
wrist-image input in Cosmos Policy. The Faster-WAM restoration shows that
this read-out carries the behavior, not that the shift arises there (9.3 to
85.3\% on Recolor+Flat, \cref{tab:alt}). Whether a repair belongs at these
sites is a separate question; repair probes answer it (\cref{fig:readout},
blue): on all three development models input-side adapters recover
0.84--0.91 of a full adapter's effect, while late-only adapters, which
contain the Faster- and Fast-WAM read-out site, recover 0.34--0.59. A late
window with the full parameter count recovers no more on Faster-WAM, so the
gap is not capacity, and every window with block 0 beats any without it.
The read-out map marks where the shift shows, not where repair works.

\section{Method: Automatic Locate-then-Repair}
\label{sec:method}

DNA mismatch repair resynthesizes a damaged strand from the intact strand
of the same molecule. Given only a released WAM, clean demonstrations and
a surface renderer, ROOT repairs it the same way and returns
$(W^\ast,t^\ast,\alpha^\ast)$: an adapter window, a checkpoint and a
deployment scale. With the interface and training profile fixed, these
follow from three pre-set thresholds $(\tau_s,\rho,\tau_d)$ on drift
coverage, relative recovery and drift ratio (values and Algorithm~S1:
Supp.~S4); no target-shift observation enters the choice.

\begin{figure}[t]
\centering
\includegraphics[width=\linewidth]{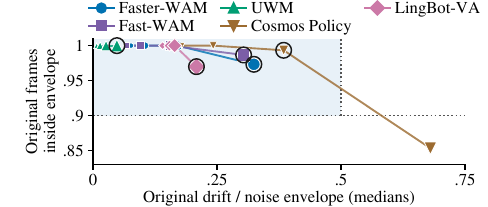}
\vspace{-1mm}
\caption{\textbf{The scale rule takes the largest $\alpha$ that keeps the original scene inside the noise envelope.} Per model, over $\alpha$: drift relative to the envelope against the fraction of original frames inside it (marker size grows with $\alpha$, rings mark $\alpha^\ast$; shaded: drift $\le0.5$, fraction $\ge0.9$).}
\label{fig:scale}
\vspace{-3mm}
\end{figure}

\paragraph{Objective: original-scene self-consistency.} The repair has one requirement: on any rendering of a frame, the
adapted model must act as the frozen model does on its original rendering. Write $v_\theta(a_\tau,\tau\mid x)$ for the
flow-matching velocity \cite{flowmatching,rectifiedflow} that the action expert predicts for a noisy action chunk $a_\tau$
at flow time $\tau$ from the video features of observation $x$; $x^0$ is a training frame in the original scene and $x^r$
the same frame under a random work-surface appearance (a \emph{random look}). The update is a low-rank adapter \cite{lora} with parameters
$\phi=\{A_\ell,B_\ell\}$: each adapted projection $W_\ell\in\mathbb{R}^{d\times k}$ of the selected window computes (\cref{eq:lora})
\begin{equation}
h=W_\ell x+\Delta W_\ell\,x,\quad \Delta W_\ell=\tfrac{\kappa}{r}B_\ell A_\ell,\quad r\ll\min(d,k),
\label{eq:lora}
\end{equation}
where $W_\ell$ is frozen, $B_\ell\in\mathbb{R}^{d\times r}$ and $A_\ell\in\mathbb{R}^{r\times k}$. The rank $r$ and the adapter
multiplier $\kappa$ are fixed by the model profile (Supp.~S4). We train $\phi$ with
\begin{equation}
\mathcal{L}(\phi)=\!\!\sum_{x\in\{x^0,x^r\}}\!\!
\big\|v_{\theta,\phi}(a_\tau,\tau\mid x)-\mathrm{sg}\big[v_\theta(a_\tau,\tau\mid x^0)\big]\big\|_2^2 ,
\label{eq:sc}
\end{equation}
with shared noise and $\tau$, where sg stops gradients through the frozen model. At $\phi=0$ the $x^0$ term vanishes and the
loss is the frozen model's own appearance gap $\lVert v_\theta(x^r)-v_\theta(x^0)\rVert^2$, so training starts from
that gap; the loss is zero exactly when the adapted model returns the frozen model's original-scene velocity on
both renderings. The template is already inside the model, so no label is needed: demonstration actions
form the noised input but are never a regression target. This is ROOT: its only target is the model's own prediction; for other action heads the matched output differs (Supp.~S4).

\paragraph{Training views and the safety envelope.} For a subset of
demonstration frames we restore the recorded state and render random
surface appearances, excluding Recolor-like hues (details: Supp.~S4);
no Recolor or Recolor+Flat rendering is ever read (an unseen color inside
a seen perturbation family). Checkpoints are selected on held-out frames
inside the \emph{original-scene safety envelope}. For held-out original frame $x^0_i$ let $e_i$ be the spread of the
native model's first action chunk under inference-noise resampling (the largest distance over four re-draws) and $\delta_i$ the
distance between the adapted and native chunks; then
\begin{equation}
\mathrm{Env}=\frac{1}{N}\sum_{i=1}^{N}\mathbf{1}\big[\delta_i\le e_i\big]\ \ge\ \tau_s
\label{eq:s1}
\end{equation}
is the fraction of held-out frames within the envelope.

\paragraph{Automatic placement and checkpoint.} Nested windows $W_1\subset\dots\subset W_K=W_{\rm full}$ start at block 0.
Each is trained briefly with \cref{eq:sc} (probe budget: Supp.~S4) and scored by Env (\cref{eq:s1}) and by \mbox{Recov} (\cref{eq:m1}), the median
held-out recovery of the original-scene first action chunk $a^0_i$ from random looks $x^{r,j}_i$ (native $a_{ij}$, adapted
$a^\phi_{ij}$):
\begin{equation}
\mathrm{Recov}=\operatorname*{median}_{i,j}\Big(1-\frac{\lVert a^{\phi}_{ij}-a^0_i\rVert}{\lVert a_{ij}-a^0_i\rVert}\Big).
\label{eq:m1}
\end{equation}
With $\mathcal A=\{k:\mathrm{Env}(W_k)\ge\tau_s,\ \mathrm{Recov}(W_k)\ge\rho\,\mathrm{Recov}(W_{\rm full})\}$ the admissible
windows, the rules choose
\begin{equation}
W^\ast=W_{\min\mathcal A},\qquad t^\ast=\operatorname*{arg\,max}_{t:\ \mathrm{Env}_t\ge\tau_s}\mathrm{Recov}_t ,
\label{eq:select}
\end{equation}
where $t$ runs over the checkpoints of the adapter trained on $W^\ast$ (fallbacks and ties: Algorithm~S1); if no window has $\mathrm{Env}\ge\tau_s$ the native model is retained, and if none reaches the $\rho$ fraction the full window (when admissible) or the highest-Recov admissible prefix is used. The read-out analysis thus supplies the design rule---act upstream of the read-out site---and the window rule sets its extent; deployment needs no privileged localization or target-shift data.

\begin{table*}[t]
\centering\footnotesize
\setlength{\tabcolsep}{2pt}
\caption{\textbf{ROOT restores shifted-surface success across WAMs while keeping the original scene.} Closed-loop success (\%) on $n$ paired fresh layouts; subscript: ROOT$-$Native (pp); bold: higher of the pair off Original; Safe: Original and Red-object non-inferior at $-10$\pp. ROOT: full window except Light-WAM (0--16); $^\S$pre-registered fallback. Complete table: Supp.~S5.1.}
\label{tab:main}
\begin{tabular*}{\textwidth}{@{\hspace{4pt}\extracolsep{\fill}}llccccccc@{\hspace{4pt}}}
\toprule
& & \multicolumn{2}{c}{Safety conditions} & \multicolumn{4}{c}{Shifted work surface} & \\
\cmidrule(lr){3-4}\cmidrule(lr){5-8}
Model ($n$) & Arm & Original\up & Red-object\up & Flat\up & Recolor\up & Recolor+Flat\up & R+F, seeds 1--3 & Safe \\
\midrule
Faster-WAM (150) & Native & 98.0 & 92.7 & 79.3 & 50.7 & 9.3 & & \\
 & ROOT & 97.3 & \textbf{96.0} & \textbf{98.0}\gain{+18.7} & \textbf{98.7}\gain{+48.0} & \textbf{96.0}\gain{+86.7} & 96.0--98.7 & \safe \\
\midrule
Fast-WAM (100) & Native & 94.0 & 75.0 & 72.0 & 65.0 & 16.0 & & \\
 & ROOT & 93.0 & \textbf{77.0} & \textbf{94.0}\gain{+22.0} & \textbf{93.0}\gain{+28.0} & \textbf{88.0}\gain{+72.0} & 88.0--90.0 & \safe \\
\midrule
UWM (150) & Native & 74.0 & 58.7 & 45.3 & 17.3 & 16.7 & & \\
 & ROOT & 70.7 & \textbf{68.7} & \textbf{70.0}\gain{+24.7} & \textbf{61.3}\gain{+44.0} & \textbf{64.0}\gain{+47.3} & 60.7--64.0 & \safe \\
\midrule
Cosmos Policy (100) & Native & 95.0 & \textbf{94.0} & \textbf{100.0} & 94.0 & 62.0 & & \\
 & ROOT & 98.0 & \textbf{94.0} & 96.0\dropd{$-$4.0} & \textbf{98.0}\gain{+4.0} & \textbf{99.0}\gain{+37.0} & 97.0--99.0 & \safe \\
\midrule
LingBot-VA (150) & Native & 99.3 & 96.7 & \textbf{96.7} & \textbf{96.0} & 90.0 & & \\
 & ROOT & 96.0 & \textbf{97.3} & 93.3\dropd{$-$3.3} & 90.0\dropd{$-$6.0} & \textbf{96.0}\gain{+6.0} & 95.3--96.0 & \safe \\
\midrule
Light-WAM$^{\S}$ (150) & Native & 93.3 & \textbf{70.0} & 0.0 & 2.7 & 0.0 & & \\
 & ROOT & 94.7 & 68.0 & \textbf{58.0}\gain{+58.0} & \textbf{76.7}\gain{+74.0} & \textbf{14.0}\gain{+14.0} & 10.0--16.0 & \safe \\
\bottomrule
\end{tabular*}
\vspace{-3mm}
\end{table*}
\paragraph{Automatic scale and deployment.}
\label{sec:dose}
The deployment scale $\alpha$ interpolates between frozen and adapted weights \cite{wortsman2022wiseft}: each adapted layer
uses $W_\ell'=W_\ell+\alpha\,\Delta W_\ell$. Like a drug titrated to its maximum tolerated dose, the rule takes the largest
$\alpha$ that stays within the model's own noise. With $\delta_i(\alpha)$ the distance between the native and $\alpha$-scaled
chunks on held-out original frame $i$ (splits: Supp.~S4),
\begin{equation}
\begin{split}
\alpha^\ast=\max\Big\{&\alpha\in\{0,\tfrac18,\tfrac14,\tfrac12,1\}:\\
&\frac{\operatorname{med}_i\delta_i(\alpha)}{\operatorname{med}_i e_i}\le\tau_d,\ \mathrm{Env}_\alpha\ge\tau_s\Big\}.
\end{split}
\label{eq:dose}
\end{equation}
\Cref{eq:dose} yields $\alpha^\ast=1$ on Faster-WAM, Fast-WAM, UWM and LingBot-VA and $\frac12$ on Cosmos Policy.
The configuration is deployed if and only if Original-scene closed-loop success passes a one-sided 95\%
non-inferiority test at $-10$\pp\ against native (Supp.~S1); otherwise
the native model is retained ($\alpha=0$).

\paragraph{What the rules guarantee.} The choice $(W^\ast,t^\ast,\alpha^\ast)$ depends only on the frozen model, the clean demonstrations and the renderer. Every input has an output, since
$\alpha=0$ is always admissible. A configuration whose true original-scene success is $10$\pp\ or more below native is
deployed with probability at most $0.05$ (asymptotically, on test-like layouts). Along $\alpha$ the drift ratio
rises and the fraction inside the envelope never increases on all five models (\cref{fig:scale}), so $\alpha^\ast$ is the boundary of
the admissible region in a line search from the noise envelope. On all five models the selection is unchanged over a wide range of each threshold, and configurations chosen outside it still repair the model in closed loop (Supp.~S4).

\paragraph{Cost.} Faster-WAM takes about 15 A100-hours with the window probes; the adapter is 0.16--1.7\% of a
WAM's parameters and can be folded into the weights of three of the models (Supp.~S8).

\section{Experiments}
\label{sec:experiments}

The experiments ask four questions; unless noted, gains over native are significant under the pre-registered paired tests of \cref{sec:setup} (Supp.~S5, S6).

\subsection{Does the repair work, and for which models?}
\label{sec:main}

\begin{figure}[t]
\centering
\includegraphics[width=\linewidth]{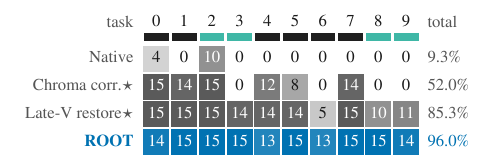}
\vspace{-5mm}
\caption{\textbf{ROOT succeeds on every task, including four chroma correction misses.} Faster-WAM, Recolor+Flat successes per task (/15); $^\star$privileged references; strip: kitchen (teal) or near-black.}
\label{fig:pertask}
\vspace{-3mm}
\end{figure}

\paragraph{ROOT restores the shifted scene without touching the original one.} On Faster-WAM, Recolor+Flat success returns to the
original-scene level (9.3 to 96.0\%) with both safety conditions non-inferior (\cref{tab:main}). The gain is largest where the model is
helpless: ROOT succeeds in nearly every episode of the eight tasks with no native success, including the four that chroma correction
leaves at zero (\cref{fig:pertask}), and it exceeds privileged restoration of the late values (\cref{tab:alt}), which indicates that
repairing at the origin does more than patching the read-out.

\paragraph{Benefit follows brittleness.} Which columns recover follows the diagnosis of \cref{sec:factors}: where lightness drives the
loss (Cosmos Policy, UWM), Recolor and Recolor+Flat recover; where hue and texture matter too (Faster-, Fast-WAM), all three
shifted columns do. The gain tracks what there is to recover, and GR00T N1.7 \cite{gr00tn1}, which loses nothing, needs no repair
(Supp.~S6.3). Original barely moves for any model: ROOT repairs what is broken and leaves the rest alone (\cref{tab:main}; Supp.~S5.1).

\paragraph{The repair acts where the shift acts.} The shift already displaces the first action chunk; ROOT pulls it back while moving
the original-scene chunk an order of magnitude less, and, for most models, its successes take as long as in the original scene
(Supp.~S5.3). The adapter thus changes perturbed actions rather than original ones.

\begin{figure*}[t]
\centering
\vfig{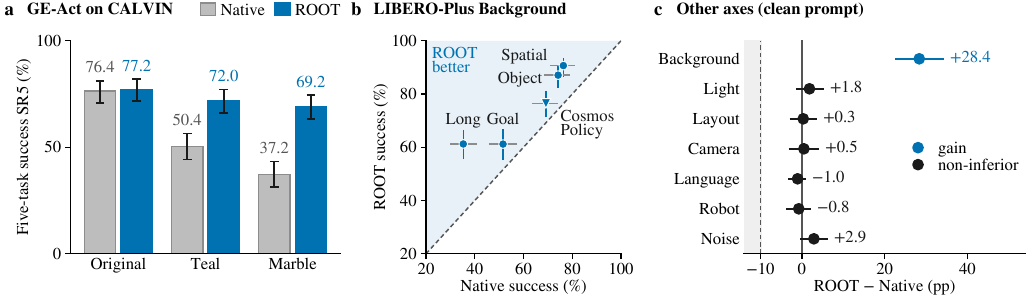}
\vspace{-1mm}
\caption{\textbf{ROOT's gains carry over to a second simulator and unseen textures.} (a)~GE-Act on CALVIN, five-task success, 250 untouched chains. (b)~LIBERO-Plus Background success, official prompt: ROOT vs.\ native, Faster-WAM on four suites, Cosmos Policy on LIBERO-10. (c)~ROOT $-$ native on other axes.}
\label{fig:boundary}
\vspace{-3mm}
\end{figure*}
\subsection{Is it just augmentation?}
\label{sec:baselines}

\begin{table}[t]
\centering\scriptsize
\setlength{\tabcolsep}{1.3pt}
\caption{\textbf{Same renderings, different targets: only ROOT recovers and passes both safety checks.} Faster-WAM, success (\%). LP-Bg.: LIBERO-Plus Background; Ctrl.: R+F when ROOT's scale rule configures the arm (arms only at $\alpha=\frac18$). $^\star$privileged reference; $^r$re-implemented.}
\label{tab:alt}
\begin{tabular*}{\columnwidth}{@{\hspace{2pt}\extracolsep{\fill}}lccccccccc@{\hspace{2pt}}}
\toprule
Arm & Orig.\up & Red.\up & Flat\up & Rec.\up & R+F\up & Safe & LP-Bg.\up & Ctrl.\up \\
\midrule
Native & 98.0 & 92.7 & 79.3 & 50.7 & 9.3 & & 35.3 & \\
\textcolor{gray}{Chroma corr.$^\star$} & \textcolor{gray}{97.3} & \textcolor{gray}{94.0} & \textcolor{gray}{81.3} & \textcolor{gray}{94.0} & \textcolor{gray}{52.0} & & \textcolor{gray}{49.8} & -- \\
\textcolor{gray}{Late-V restore$^\star$} & \textcolor{gray}{97.3} & \textcolor{gray}{96.0} & \textcolor{gray}{96.0} & \textcolor{gray}{94.0} & \textcolor{gray}{85.3} & & -- & -- \\
\midrule
Aug-BC & 92.7 & 81.3 & 84.0 & 94.0 & 85.3 & \unsafe & 42.2 & 26.0 \\
Input-pert.$^{r}$ & 91.3 & 84.7 & 90.7 & 93.3 & \underline{86.7} & \unsafe & \underline{43.9} & \underline{32.7} \\
Rep.-anchor.$^{r}$ & 87.3 & 76.7 & 84.0 & 88.7 & 76.0 & \unsafe & 40.1 & 27.3 \\
CLAHE & 98.7 & 92.7 & 79.3 & 66.7 & 7.3 & \safe & -- & -- \\
Gray-world & 97.3 & 92.0 & 50.7 & 60.0 & 11.3 & \safe & -- & -- \\
Steering (late) & 98.7 & 90.7 & 83.3 & 54.0 & 14.0 & \safe & -- & -- \\
\midrule
ROOT, 0--7 & 96.0 & \underline{94.0} & \underline{96.7} & \underline{97.3} & \textbf{96.0} & \safe & -- & -- \\
ROOT & 97.3 & \textbf{96.0} & \textbf{98.0} & \textbf{98.7} & \textbf{96.0} & \safe & \textbf{61.2} & \textbf{96.0} \\
\bottomrule
\end{tabular*}
\vspace{-3mm}
\end{table}

\paragraph{The target, not the renderings, makes the repair safe.} Arms supervised with demonstration actions on the same renderings
(Aug-BC, RobustVLA's input branch \cite{robustvla} and representation anchoring \cite{anchoralign}, the last two re-implemented) recover
much of Recolor+Flat but fail Original or Red-object non-inferiority: original-scene actions leave the model's noise envelope far more
often than under ROOT, because labels steer the policy toward the demonstrator, whereas ROOT's clean-view target keeps the original scene
fixed. Under ROOT's scale rule these arms deploy only at one eighth of their update, which removes most of the recovery (Ctrl.,
\cref{tab:alt}). In contrast, preprocessing \cite{zuiderveld1994clahe,buchsbaum1980} and steering leave Recolor+Flat close to native, as
the diagnosis predicts: image statistics fix color but not texture, and the read-out site is not where repair works
(\cref{sec:chroma,sec:locate}).

\subsection{What each part contributes}
\label{sec:ablate}

\begin{table}[t]
\centering\scriptsize
\setlength{\tabcolsep}{2pt}
\caption{\textbf{Anchor, window and scale each carry recovery.} Faster-WAM, Recolor+Flat success (R+F, \%) on 150 further layouts; each row changes one thing from ROOT, paired with a same-process ROOT re-run. $\Delta$: R+F $-$ ROOT, pp. All rows keep Original and Red-object non-inferior (Supp.~S5.2).}
\label{tab:ablate}
\begin{tabular*}{\columnwidth}{@{\hspace{2pt}\extracolsep{\fill}}lcc@{\hspace{2pt}}}
\toprule
Variant & R+F & $\Delta$ R+F \\
\midrule
ROOT & 98.7 & -- \\
no clean-view ($x^0$) term & 92.0 & $-6.7$ \\
blocks 16--28 instead of 0--7 & 85.3 & $-13.3$ \\
rank 8 instead of 32 & 94.7 & $-4.0$ \\
scale $\frac18$ instead of 1 & 30.7 & $-67.3$ \\
\bottomrule
\end{tabular*}
\vspace{-3mm}
\end{table}

\paragraph{The anchor holds the original scene; window and scale carry the repair.} Without the clean-view term only one of 16 checkpoints
keeps original-scene actions inside the model's noise envelope, and with an added demonstration-action loss none does (all 16 do for
ROOT). The other ablations cost recovery but not safety (\cref{tab:ablate}; Supp.~S5.2): one eighth of the update removes most of the
gain, the late window, the read-out site, costs the next most, and a smaller rank costs little.

\begin{figure*}[t]
\centering
\input{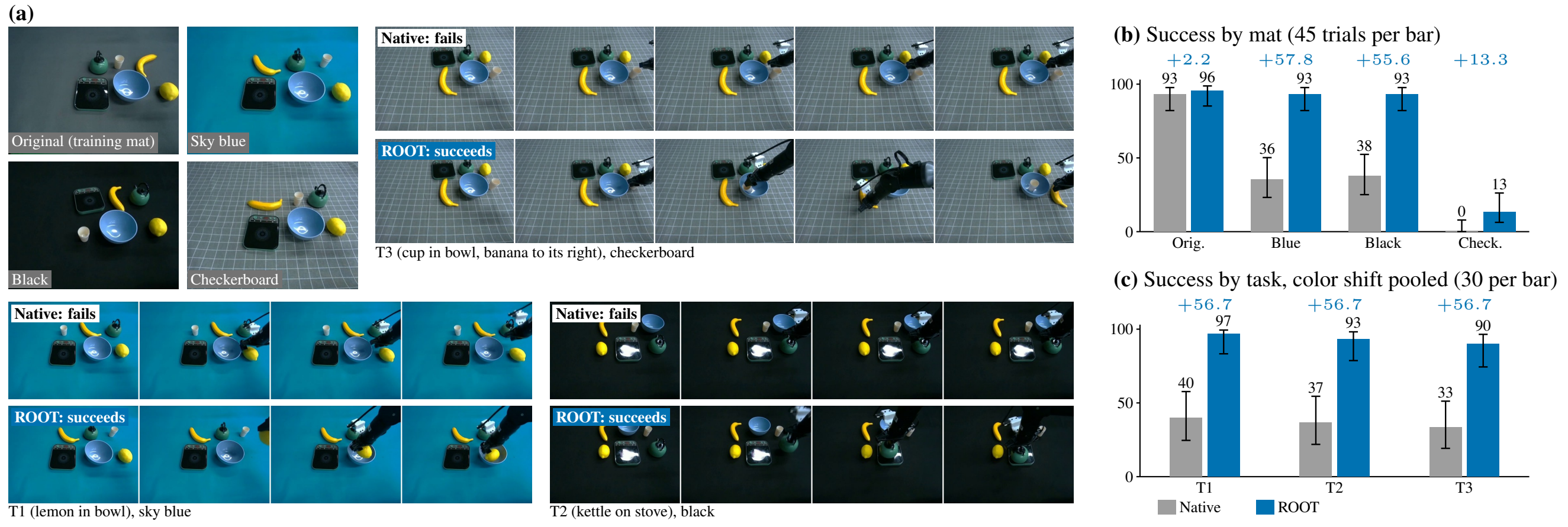}
\vspace{-1mm}
\caption{\textbf{ROOT succeeds on a real robot.} (a)~The four mats (gray: from the training demonstrations) and third-person rollouts of both models across mats and tasks. (b)~Success per mat. (c)~Success per task; all counts in Supp.~S7.}
\label{fig:realrobot}
\vspace{-3mm}
\end{figure*}
\subsection{Where to repair, and who decides}
\label{sec:placement}

\paragraph{A narrow early window suffices on Faster- and Fast-WAM.} The rule selects the first blocks on the Wan2.2 models, and the
narrow window is non-inferior to the full adapter on Fast-WAM in every seed and on Faster-WAM in a confirmatory test (Supp.~S5.4); on
Cosmos Policy blocks 0--13 reach 98\% Recolor+Flat success against 99\% for the full adapter (Supp.~S5.4).

\paragraph{A screen that acts.} On DiT4DiT, whose native model fails Recolor and Recolor+Flat outright, early windows still recover the
most, but none keeps 90\% of original-scene drifts inside the envelope at full strength, so the controller keeps the native model. On
Light-WAM no window passes either, so the pre-registered fallback applies the scale rule one stage earlier and deploys an early window at
half strength, with Original and Red-object non-inferior (\cref{tab:main}). The screen thus acts before the scale does (Supp.~S5.4).

\paragraph{The controller can run alone.} With controller, thresholds and data splits frozen, we let it configure Faster-WAM and
Streaming-WAM \cite{streamingwam}, a checkpoint unused in development (three seeds each). All six runs deployed; Faster-WAM again
reaches near-perfect Recolor+Flat success and Streaming-WAM rises from 27.3 to 83.1\%, with all Recolor+Flat and safety tests passing
(Supp.~S5.4). The rules thus transfer to a model they were never tuned on.

\section{Beyond SurfaceShift}
\label{sec:boundaries}

\paragraph{The mechanism carries to another simulator and model family.} GE-Act \cite{genieenvisioner} predicts actions from an LTX-style video
transformer and is evaluated on CALVIN \cite{calvin} as chains of five language-specified tasks. Recoloring the wooden
table in the render path (segmentation-guided, physics untouched) lowers its five-task success for a teal table and for an unseen marble texture
(\cref{fig:boundary}a). With only the adapter and re-render interface ported, the frozen
controller chose the early blocks at full strength, trained on the model's own clean-scene rollouts, and deployed: ROOT restores
both tables and the original scene stays non-inferior (Supp.~S6). The procedure is thus not tied to LIBERO or to the backbones it was developed on.

\paragraph{The repair transfers to textures it has never seen.} LIBERO-Plus \cite{liberoplus} replaces the table with 136 textures absent from
ROOT's training. Background success rises from 35.3 to 61.2\% under the official prompt (\cref{fig:boundary}b), to our knowledge the largest
Background gain reported for a post-hoc repair without architecture change or LIBERO-Plus data. Which variants ROOT
rescues is predicted by texture rather than color (\cref{fig:factors}b), the component chroma correction cannot repair; the same-budget
arms of \cref{tab:alt} gain far less and fall below native on other axes, where ROOT stays non-inferior (Supp.~S6.2).

\paragraph{It stays inert where it was not built to act.} Run as is on the Spatial, Object and Goal suites, the Faster-WAM adapter trained
on LIBERO-10 only raises Background on all three, and Cosmos Policy gains as well (\cref{fig:boundary}b). On the other axes ROOT stays
non-inferior on the clean prompt (\cref{fig:boundary}c; Supp.~S6.2), consistent with a repair specific to surface appearance; Lighting is
future work.

\paragraph{A real robot confirms the repair and its reach.}
A Piper arm runs three tasks on four mats. Faster-WAM is fine-tuned on
120 teleoperated demonstrations from the gray mat (40 per task, object
layouts randomized during collection and mixed in training;
\cref{fig:realrobot}a; Supp.~S7). ROOT uses in-image recoloring in place
of a renderer; every other choice is made by the frozen rules. The
unseen sky-blue and black mats keep the trained instructions but
rearrange the surrounding objects. The repair generalizes across these
placements, and the mat color remains the only shift. With matched
placements and a blinded operator, ROOT restores native's 36.7\% success
to 93.3\% on every task (\cref{fig:realrobot}b,c).

On the gray mat ROOT is non-inferior. The checkerboard mat has a
high-contrast texture unseen in training. Even there, ROOT lifts success
from zero (6 of 45 trials); native never succeeds.

\section{Conclusion and Discussion}
\label{sec:conclusion}

Changing only the table's color and texture drops a
near-perfect WAM from 98.0 to 9.3\% success. This failure should not be
repaired where it shows: previous direct repair comes from the input-side
blocks alone, not from the late ones. ROOT turns this finding into a
practical repair. It is label-free: the adapter learns from the model's
own original-scene predictions, with no target-shift data. It is
automatic: fixed rules choose where to adapt and how strongly to deploy.
It is safe: it deploys only if original-scene success is preserved. It
is cheap: the adapter is a small fraction of the model's parameters.
Across nine WAMs, two simulators and a real robot, the gains are large
and general. The same frozen rules work on a model unused in development
and in a second simulator, and the repair transfers to unseen
LIBERO-Plus textures and to a real robot.

Discussion. Our experiments point to three next steps.
\emph{Texture}: texture views already lift held-out textures from 100 to
143 of 150 in simulation (Supp.~S5.2); patterned-cloth recoloring in the
real-robot pairs is the natural way to bring this to the robot.
\emph{Illumination}: lighting views are not yet in training; they would
extend the same objective unchanged. \emph{Supervision and validation}:
the training pairs need one state under two surfaces, which a simulator
or segmentation-guided recoloring supplies; training on the model's own
rollouts already removes the demonstrations (Supp.~S5.2), propagated masks
could remove the segmentation, and the offline envelope fraction
separates safe from unsafe repairs without closed-loop trials
(Supp.~S5.3), a route to a rollout-free check.

\section*{LLM Usage Statement}
We used large language models throughout this work. Anthropic's Claude (through the Claude Code assistant) helped us with the writing. OpenAI's Codex wrote part
of the code, including the configuration controller. The authors specified the research questions, methods, experimental designs
and decision rules, reviewed the results, and took responsibility for all content.

{\small
\bibliographystyle{ieeenat_fullname}
\bibliography{refs}
}

\clearpage
\onecolumn
\setcounter{section}{0}\setcounter{table}{0}\setcounter{figure}{0}\setcounter{algorithm}{0}
\renewcommand{\thesection}{S\arabic{section}}
\renewcommand{\thetable}{S\arabic{table}}
\renewcommand{\thefigure}{S\arabic{figure}}
\renewcommand{\thealgorithm}{S\arabic{algorithm}}
\renewcommand{\theHsection}{S\arabic{section}}
\renewcommand{\theHtable}{S\arabic{table}}
\renewcommand{\theHfigure}{S\arabic{figure}}
\renewcommand{\theHalgorithm}{S\arabic{algorithm}}
\setlist{leftmargin=*,itemsep=1pt,topsep=2pt}
\renewcommand{\topfraction}{0.95}\renewcommand{\bottomfraction}{0.9}\renewcommand{\textfraction}{0.05}\renewcommand{\floatpagefraction}{0.6}
\setlength{\textfloatsep}{8pt plus 2pt minus 2pt}\setlength{\floatsep}{6pt plus 2pt minus 2pt}\setlength{\intextsep}{6pt plus 2pt minus 2pt}\setlength{\abovecaptionskip}{3pt}\setlength{\belowcaptionskip}{0pt}
\setcounter{topnumber}{4}\setcounter{bottomnumber}{3}\setcounter{totalnumber}{7}
\providecommand{\ph}{}
\newcommand{\off}{\textsuperscript{$\diamond$}}
\newcommand{\pv}{\textsuperscript{$\star$}}
\newcommand{\xr}[1]{\ref{#1}}
\newcommand{\xeqref}[1]{\textup{(\ref{#1})}}
\begin{center}{\large\bfseries Label-free Self-repair of Visual Fragility in World Action Models\\[2pt] Supplementary Material}\end{center}\vspace{-4pt}
\noindent This supplement follows the order of the main paper; numbers prefixed with S refer to it, all others to the main paper.

\paragraph{Evidence.} Closed-loop numbers are paired rollouts on fresh layouts unless a table names another state set
(diagnostic sets, LIBERO-Plus variants, CALVIN chains, 400-step offline probes). Closed-loop protocols, checkpoints and
tests were frozen before their runs. Markers: $^\diamond$offline action test; $^\star$privileged
reference that reads same-state original-scene information. LIBERO-Plus entries are paired changes over the same base
(S6.2).

\section{Conditions, Models and Protocol}
\label{sec:s-setup}

\subsection{Conditions}

\begin{table}[!htb]
\centering\scriptsize
\setlength{\tabcolsep}{3.5pt}
\caption{\textbf{SurfaceShift conditions and their rendered effect.} (a) Only rendering changes; physics, object poses and
camera equal Original, and so does the instruction except in the Red-object conditions. Recolor multiplier:
RGB $[0.08, 0.45, 0.72]$. (b) Medians over 15 layouts per scene; $\Delta E_{00}$: CIEDE2000 to Original. Flat is a
pure texture change only in the kitchens.}
\label{tab:s-conditions}
\textbf{(a) Conditions}\\[2pt]
\begin{tabular}{@{}lll@{}}
\toprule
Name & Intervention & Role \\
\midrule
Original & released scene & safety \\
Flat & texture $\to$ per-channel mean & target \\
Recolor (R) & fixed blue multiplier on the material, texture kept & target \\
Recolor+Flat (R+F) & multiplier and flattening & primary target \\
Red-object & target object red; instruction says ``red'' & safety \\
Red-object, neutral & target object red; neutral instruction & extension \\
Red-object, conflicting & target object red; wrong color in instruction & extension \\
\bottomrule
\end{tabular}

\medskip
\textbf{(b) Rendered effect on the work surface}\\[2pt]
\begin{tabular}{@{}l rrr rrr rrr@{}}
\toprule
& \multicolumn{3}{c}{kitchen (tasks 2, 3, 8, 9)} & \multicolumn{3}{c}{living room (0, 1, 4, 6, 7)} & \multicolumn{3}{c}{study (5)} \\
\cmidrule(lr){2-4}\cmidrule(lr){5-7}\cmidrule(lr){8-10}
Measure & Flat & R & R+F & Flat & R & R+F & Flat & R & R+F \\
\midrule
$\Delta E_{00}$ & 0.5 & 28.2 & 28.2 & 6.9 & 23.0 & 20.7 & 21.0 & 30.4 & 29.6 \\
$\Delta L$ & $-0.2$ & $-8.7$ & $-8.9$ & $-9.2$ & $-16.3$ & $-19.3$ & $-24.4$ & $-23.4$ & $-32.0$ \\
$|\Delta H|$ & 0.4 & 31.4 & 31.3 & 0.1 & 19.4 & 13.9 & 0.5 & 27.8 & 16.9 \\
$\Delta\log$ texture & $-3.67$ & 0.03 & $-1.91$ & $-1.84$ & $-0.26$ & $-1.75$ & $-1.80$ & 0.05 & $-1.56$ \\
\bottomrule
\end{tabular}
\end{table}

In the kitchen tasks the lit material saturates in green and blue, so the multiplier mainly removes about 58\% of the red
channel: bright teal, CIE $L^*a^*b^*\approx(65,-31,-2)$ against $(74,3,12)$ in Original. In the living-room and study tasks
nothing saturates and the table renders near-black blue, $\approx(3.6,-2.3,-3.1)$ and $(3.2,-2.3,-2.4)$, against $(23,8,10)$
and $(39,8,26)$. Tasks are numbered 0--9 in LIBERO-10's order.

\subsection{Models, layouts and statistics}
\label{sec:s-stat}

\emph{Contracts.} All models are released checkpoints except UWM (our fine-tune, Sec.~\ref{sec:s-instantiations}). Faster-WAM,
Fast-WAM and Light-WAM: horizon 700, 30 settle steps, re-planning every 10 steps; Cosmos Policy: 520 steps after 10 settle
steps; UWM: $128^2$ rendering, 0 settle steps, 1000 steps, two-frame history. Faster- and Fast-WAM re-seed their fixed
inference noise at every query, so trial seeds are pseudo-replicates: the unit is the (task, layout) pair, with a second noise
seed as replication. \emph{Layouts.} Fresh layouts come from LIBERO's BDDL sampler, seeded deterministically, and are
accepted only if they restore bit-exactly, have no interpenetration, settle within tolerances calibrated on the official
states, are not already solved and differ from every official and accepted layout; the largest displacement of any free
object from the closest official state is 0.31--2.28\,cm per task. The first 15 per task form the formal set (Fast-WAM and
Cosmos Policy: first 10). \emph{Statistics.} With $b$ pairs favoring the new arm and $c$ the reference, the one-sided exact
sign test has $p=\sum_{k=b}^{b+c}\binom{b+c}{k}2^{-(b+c)}$ (1 if $b+c=0$). Non-inferiority at $-10$\pp\ is a one-sided 95\% test; tables report its lower bound (LB), from the paired Tango score test for
per-seed narrow-window comparisons, LIBERO-Plus and the controller-run models, from a pair-level bootstrap (10{,}000 resamples)
in the main SurfaceShift runs, and from a layout-cluster bootstrap for pooled narrow-window and
inference-seed tests.

\section{SurfaceShift: Protocol and Datasheet}
\label{sec:s-surfaceshift}

SurfaceShift is a protocol for work-surface appearance shifts: it asks which appearance axis hurts which model and whether a
repair changes original-scene behavior. It contains the conditions of Table~\ref{tab:s-conditions}; 112 appearance cells
written into the surface texture (a lightness grid $L^*$ 5--85 crossed with gray and six hues, a low-chroma set, recolored
original and three fabric textures at five lightness levels, the bright-teal and near-black looks on all ten tasks, the
original conditions as anchors and a zero-color-shift texture ladder; fractional, not full factorial); 15 fresh layouts per
task (150 pairing keys); and native per-episode outcomes of eight policies (Table~\ref{tab:s-main-full}). Cells are calibrated per scene and channel from an 11-level gray probe and analyzed at
their measured $L^*a^*b^*$ (all 132 flat (cell, scene) pairs within 5 $\Delta E_{00}$ of target; the living-room and study
surfaces cannot exceed $L^*$ 76.5 and 69.8). Each trial creates a fresh environment and sets the paired seed before the reset;
GPU rendering is not bit-reproducible, so paired comparisons run within one process on one hardware type. The scope is
appearance of the LIBERO-10 work surface (geometry, camera, language, lighting and other suites are covered by LIBERO-Plus).
The release will hold generators, cell tables, layout keys and seed rule, native-only evaluation adapters with pinned upstream
commits, statistics and envelope code, the configuration controller and native outcomes (code MIT, data CC BY 4.0, to be
confirmed), and part of the training code (ROOT on Faster-WAM); all code will be released upon publication. Model weights and
LIBERO-Plus textures are not redistributed.

\section{Diagnosis Details}
\label{sec:s-diagnosis}

\subsection{Native collapse and chroma correction}

\begin{table}[!htb]
\centering\scriptsize
\setlength{\tabcolsep}{3pt}
\caption{\textbf{Recolor+Flat collapses native success; chroma correction repairs color but not texture.} Paired matrices
(10 tasks $\times$ 50 official states, successes /500; partial rollouts, each under its own horizon), native policies
re-measured under the official contracts (pre-registered), and chroma correction (native $\to$ corrected, official-contract
states; counts out of the contract's denominator, 200 or 80). $^a$copy-frame inference, $256^2$, 10 settle steps;
$^b$environment steps (796 policy steps).}
\label{tab:s-phenomenon}
\label{tab:s-chroma}
\begin{tabular}{@{}l r rrrr l rrr rr@{}}
\toprule
& & \multicolumn{4}{c}{Matrices (/500)} & & \multicolumn{3}{c}{Official contract} & \multicolumn{2}{c}{Chroma correction} \\
\cmidrule(lr){3-6}\cmidrule(lr){8-10}\cmidrule(lr){11-12}
Model & Horizon & Orig. & Flat & Recolor & R+F & tasks$\times$states & Orig. & Recolor & R+F & Recolor & R+F \\
\midrule
Faster-WAM & 400 & 476 & 386 & 207 & 46 & 10$\times$20 & 190/200 & 88/200 & 23/200 & 88$\to$184 & 23$\to$93 \\
Fast-WAM & 400 & 445 & 337 & 269 & 44 & 10$\times$20 & 183/200 & 115/200 & 20/200 & 115$\to$173 & 20$\to$88 \\
Cosmos Policy & 520+10 & 489 & 481 & 461 & 351 & 4$\times$20 & 79/80 & 72/80 & 47/80 & 72$\to$77 & 47$\to$70 \\
UWM & 700$^a$ & 413 & 285 & 117 & 87 & 4$\times$20 & 46/80 & 14/80 & 10/80 & 14$\to$6 & 10$\to$6 \\
LingBot-VA & $\sim$800$^b$ & 486 & 479 & 473 & 442 & --- & --- & --- & --- & --- & --- \\
\bottomrule
\end{tabular}
\end{table}

Chroma correction converts each camera image to CIE $L^*a^*b^*$ (D65), shifts the spatial means of $a^*$ and $b^*$ to per-camera
references (means over 8 frames of each of the 500 training demonstrations: agent view $[39.61, 3.47, 7.64]$, wrist
$[40.71, 4.49, 8.79]$), keeps $L^*$ and converts back; it has no trainable parameters. On Faster-WAM (10 tasks $\times$ 5 states,
full first action chunk) it recovers 0.413 [0.302, 0.535] of the Recolor+Flat deviation, 0.615 [0.572, 0.649] of the Recolor
deviation and 0.003 [$-0.005$, 0.011] of the Flat deviation\off.

\subsection{Appearance factors and sweep}
\label{sec:s-af}

For the first frame of every evaluated episode we measured the change in lightness $\Delta L$, hue $\Delta H$, chroma $\Delta C$
and a Weber texture measure against the matched Original frame (Table~\ref{tab:s-conditions}b). Effects are average marginal
effects (AME) of logistic regressions with all axes entered jointly and task fixed effects; intervals are cluster-bootstrap 95\%
(2{,}000 resamples; clusters: layouts, or texture files on LIBERO-Plus); Holm within each family.

\begin{table}[!htb]
\centering\scriptsize
\setlength{\tabcolsep}{3pt}
\caption{\textbf{Different axes predict native failure in different models; on LIBERO-Plus, rescue depends on texture alone.}
AME in pp [95\% CI] (Holm). (a) Original, Flat, Recolor and Recolor+Flat, within task; per 10 units of lightness or hue change
and per log unit of texture change; 30-test family; LingBot-VA from per-task counts. (b) LIBERO-Plus Background and Light
(Faster-WAM, clean prompt, 563 variants; per SD; 8-test family): native success and rescue (ROOT succeeds where native failed;
195 native failures). Hue and chroma are collinear (VIF 4.2).}
\label{tab:s-af-c}
\label{tab:s-af-lp}
\textbf{(a) Native failure by axis, five models}\\[2pt]
\begin{tabular}{@{}lccccc@{}}
\toprule
Axis & Faster-WAM & Fast-WAM & UWM & Cosmos Policy\ph & LingBot-VA \\
\midrule
lightness & $-21.9$ [$-25.6$, $-18.1$] & $-10.6$ [$-16.7$, $-4.1$] & $-26.8$ [$-33.3$, $-20.6$] & $-15.8$ [$-20.8$, $-11.5$] & $-0.2$ [$-2.3$, $+1.9$] \\
hue & $-14.8$ [$-18.4$, $-11.7$] & $-16.0$ [$-18.8$, $-12.7$] & $-2.6$ [$-5.9$, $+1.0$] & $-0.8$ [$-5.4$, $+3.8$] & $-2.8$ [$-4.2$, $-1.5$] \\
texture & $-12.7$ [$-16.3$, $-9.6$] & $-14.4$ [$-17.5$, $-11.6$] & $-0.4$ [$-2.8$, $+1.9$] & $-2.9$ [$-8.0$, $+1.2$] & $-2.3$ [$-3.9$, $-0.9$] \\
\bottomrule
\end{tabular}

\medskip
\textbf{(b) LIBERO-Plus, Faster-WAM}\\[2pt]
\begin{tabular}{@{}lcccc@{}}
\toprule
 & lightness (SD 14.8) & hue (SD 15.3) & chroma (SD 13.6) & texture (SD 0.76) \\
\midrule
native success (Holm) & $-4.4$ [$-7.2$, $-1.4$] (0.012) & $-7.3$ [$-13.3$, $-2.1$] (0.040) & $-10.8$ [$-17.4$, $-3.1$] (0.030) & $-7.1$ [$-10.1$, $-4.0$] (0.004) \\
rescue (Holm) & $+1.0$ [$-4.5$, $+6.6$] (1.0) & $-1.4$ [$-10.8$, $+8.0$] (1.0) & $+3.8$ [$-2.9$, $+13.2$] (0.73) & $-16.9$ [$-21.4$, $-10.4$] (0.004) \\
\bottomrule
\end{tabular}
\end{table}

\paragraph{Pre-registered sweep.}\label{sec:s-sweep} No model had been run on any of the 112 sweep cells (Sec.~\ref{sec:s-surfaceshift})
when their calibration was frozen. Offline: first-chunk action tests on the 50 official states; $D$ is the native action
deviation over the native noise envelope, $R$ the fraction of it recovered by the repair (defined for $D>2$). Closed loop:
Faster-WAM, UWM and Cosmos Policy on fresh layouts 1--10, 1{,}700 episodes per model.

\begin{table}[!htb]
\centering\scriptsize
\setlength{\tabcolsep}{2.5pt}
\caption{\textbf{Eight of nine primary hypotheses of the appearance sweep are rejected} (one-sided, Holm at 0.05). C: closed
loop, O: offline\off; counts in parentheses are discordant pairs. Kitchen tasks: 2, 3, 8, 9 (layouts 1--10, 40 pairs).
In-range color cells: flat cells within 5 $\Delta E_{00}$ of the scene's rendered training appearances. Brightening cell: in
each darker scene, the gray flat cell whose $|\Delta L|$ is closest to that of the darkest gray cell ($L^*{=}5$).}
\label{tab:s-sweep}
\begin{tabular}{@{}l l p{0.4\linewidth} l l c@{}}
\toprule
& Stage, model & Hypothesis & Result & Holm $p$ & rejected \\
\midrule
H1a & C, Cosmos Policy & near-black look on kitchen tasks lowers native success vs.\ Original & 38/40 $\to$ 0/40 (38/0) & $2.9\times10^{-11}$ & yes \\
H1b & C, UWM & same & 35/40 $\to$ 0/40 (35/0) & $2.0\times10^{-10}$ & yes \\
H2a & O, Cosmos Policy & $\log D$: lightness slope $>$ hue slope (flat cells) & $b_L-b_H=0.126$ [0.089, 0.164] & 0.0015 & yes \\
H2b & O, UWM & same & $0.160$ [0.097, 0.226] & 0.0015 & yes \\
H3a & O, Faster-WAM & $R$ on in-range color cells $>$ $R$ on high-energy texture at original color & 46 vs.\ 0 states & $1.3\times10^{-13}$ & yes \\
H3b & C, Faster-WAM & rescue on that texture $<$ rescue on near-black and bright-teal looks & 3/14 vs.\ 171/187 & $5.4\times10^{-8}$ & yes \\
H4a & O, Faster-WAM & darker scenes: $R$(brightening cell) $<$ $R$(darkest gray) & 28 vs.\ 1 states & $2.8\times10^{-7}$ & yes \\
H4b & O, Fast-WAM & same & 25 vs.\ 1 states & $1.6\times10^{-6}$ & yes \\
H4c & O, UWM & same & 1 vs.\ 0 states & 0.50 & no \\
\bottomrule
\end{tabular}
\end{table}

\subsection{Read-out sites}

\begin{table}[!htb]
\centering\scriptsize
\caption{\textbf{The shift is read out at different sites in different models.} Main offline read-out site of the Recolor+Flat
shift per model (privileged same-state restoration; offline first-chunk recovery)\off\pv; denominators and action scales differ
across rows.}
\label{tab:s-readout}
\begin{tabular}{@{}lll@{}}
\toprule
Model & Main read-out site & Recovery (controls) \\
\midrule
Faster-WAM & late video values (layers 16--28) & 0.656 (early 0.004, middle 0.018) \\
Fast-WAM & late video keys/values & 0.561, 3 tasks (early 0.067) \\
UWM & early global conditioning / gate & 0.411 / 0.497 (late 0.000) \\
Cosmos Policy & wrist-camera input (pixels) & 0.654 (agent view $-0.032$; latest probed block 0.27) \\
LingBot-VA & middle keys/values (layers 10--19) & 0.635 (early $\approx$0.05) \\
\bottomrule
\end{tabular}
\end{table}

Closed-loop late-value restoration on four Faster-WAM tasks (0, 1, 3, 4) raises Recolor+Flat from 11/80
to 78/80 (Original native 78/80); on fresh layouts it reaches 144/150 on Flat, 141/150 on Recolor and 128/150 on Recolor+Flat\pv.

\section{Method Details}
\label{sec:s-method}

\subsection{Instantiations}\label{sec:s-instantiations}

\paragraph{Faster-WAM and Fast-WAM.} Every fourth training frame is rendered under $K=4$ random appearances (HSV color
$H\in[0,1)$, $S\in[0.2,0.9]$, $V\in[0.3,0.9]$ multiplied into the texture, flattened with probability 0.5); views within
$\Delta H<0.05$ of the Recolor hue are excluded. With five demonstrations per task held out, training
has 31{,}148 frames and 112{,}532 admissible views (12{,}428 excluded). Eq.~\xeqref{eq:sc} has no video loss. Rank-32 LoRA on the self-attention
and feed-forward projections of the video expert, blocks 0--28 (Faster-WAM, 55.1M parameters) and 0--29 (Fast-WAM, 57.0M);
action expert frozen. Every 1{,}500 steps, 150 held-out frames with 299 random-appearance pairs select the step with the
largest median recovery subject to Env $\ge0.90$: Faster-WAM step 16{,}500 (recovery 0.578, Env 0.973), Fast-WAM step 18{,}000
(0.561, 0.967), Faster-WAM second seed step 23{,}400 (0.588, 0.96). The other ports differ in adapter site and matched output
(Table~\ref{tab:s-ports}); all deploy at scale 1 except Cosmos Policy ($\frac12$). UWM is our fine-tune; its adapter repeats the current frame as both history frames.

\begin{table}[!htb]
\centering\scriptsize
\setlength{\tabcolsep}{3pt}
\caption{\textbf{Instantiations on the three other models.} Held-out recovery of the selected checkpoint in brackets.}
\label{tab:s-ports}
\begin{tabular}{@{}l p{0.2\linewidth} p{0.37\linewidth} p{0.28\linewidth}@{}}
\toprule
Model & Adapter & Matched output & Training; deployed \\
\midrule
UWM & rank-16 LoRA on every convolution (rank 32 on the projection) of the image encoder & student action-token noise prediction on original and random views vs.\ frozen teacher on the original view; shared noised action, DDIM step (one of ten) and future-latent noise & 6{,}000 steps on 31{,}237 frames, no no-op filter (0.686); step 6{,}000 \\
Cosmos Policy & rank-32 LoRA, all 28 blocks, image-condition tokens only & decoded-action error $+\,0.1\times$ action-slot latent error (all rows) $+\,0.1\times$ summed latent error of future proprioception, image and value slots (original rows only) & 9{,}000 steps on 27{,}834 frames; step 9{,}000, scale $\frac12$ (envelope fraction 0.953; 0.867 at $\frac34$) \\
LingBot-VA & rank-16 LoRA, self-attention of all 30 blocks (video tokens) and video patch embedding & video and action outputs of the frozen teacher on the same noised tokens & random views rendered at every step (450 demonstrations, two appearances per episode); step 4{,}500 of 6{,}000 (0.248, Env 1.00) \\
\bottomrule
\end{tabular}
\end{table}

\subsection{Procedure and properties}\label{sec:s-placement-proc}\label{sec:s-rule-props}

The probes cover contiguous block ranges, and the controller selects only among the nested prefixes that start at block 0 (Faster-WAM: 0--7, 0--15, 0--28); the other ranges are diagnostic probes. Faster-WAM (29 blocks): 0--7, 8--15, 16--28, 0--15, 8--28, 0--28 at rank 32,
plus capacity controls (0--7 at rank 112, 16--28 at rank 64; not eligible prefixes); Fast-WAM: the same over 30 blocks; Cosmos
Policy: 0--13, 14--27, 0--27; DiT4DiT: 0--5, 6--11, 12--17, 0--11, 6--17, 0--17. Held-out frames: three per held-out
demonstration (at $0,\frac13,\frac23$ of its length), two random looks each; $e_i$ is the maximum over four inference-noise
re-draws (seeds 196--199); only the median pair recovery is used (denominators floored at $10^{-12}$). The drift ratio uses a separate \emph{rule hold-out} that no selection used (Cosmos Policy: the selection
hold-out); if no $\alpha$ qualifies, the procedure stops. Ties favor the smaller window and the earliest checkpoint. Light-WAM profile: if no window passes at $\alpha=1$, the probes are re-read at $\alpha=\frac12$ (pre-registered branch; the Faster- and Streaming-WAM runs have none). The
controller reads one verdict from the frozen Original evaluator, a one-sided 95\% test at $-10$\pp\ (paired Tango score test;
Sec.~\ref{sec:s-stat}), run once on the selected configuration and on layouts no
selection step used, so there is no multiplicity over $\alpha$, windows or checkpoints.

\begin{table}[!htb]
\centering\scriptsize
\caption{\textbf{Fixed hyperparameters and automatically selected outputs.} The first two rows are the Faster-/Fast-WAM
profile; thresholds are shared by all models.}
\label{tab:s-auto-hyper}
\begin{tabular}{@{}p{0.24\linewidth}p{0.37\linewidth}p{0.31\linewidth}@{}}
\toprule
Quantity & Fixed value or profile & Role \\
\midrule
LoRA / optimizer & $r=32$, $\kappa=32$; AdamW, $\eta=10^{-4}$ & Fixed training configuration \\
Training budget & 23{,}400 steps; 200 warm-up; batch $4\times2$ & Checkpoint candidates on fixed schedule \\
Probe budget & $B=400$ steps per window & Compare eligible prefixes \\
Coverage / recovery & $\tau_s=0.90$, $\rho=0.80$ & Select $W^\ast$, then $t^\ast$ \\
Scale constraint & $\tau_d=0.50$, $\mathcal A=\{1,\frac12,\frac14,\frac18\}$ & Select largest admissible $\alpha^\ast$ \\
Original preservation & NI margin $-10$\pp; frozen test & Deploy or retain native weights \\
Automatic outputs & $W^\ast$, $t^\ast$, $\alpha^\ast$ & No target-shift tuning \\
\bottomrule
\end{tabular}
\end{table}

\begin{algorithm}[h]
\caption{ROOT automatic configuration under a fixed model profile.
Inputs are clean demonstrations and random surface appearances; no target-shift observations or action targets.}
\label{alg:ltr}
\small
\begin{algorithmic}[1]
\Require frozen WAM $f_\theta$, clean demos $\mathcal D$, surface renderer, fixed profile $h$
\State Generate 4 random surface looks per 4th frame; estimate native noise envelope $E$
\State Probe eligible prefixes for $B=400$ steps; measure \mbox{Recov} and Env
\State $\mathcal S\gets\{W:\mathrm{Env}(W)\ge\tau_s\}$; if empty, \Return native (Light-WAM profile: first re-probe at $\alpha=\tfrac12$)
\State If $\mathrm{Recov}_{\rm full}>0$, choose the smallest prefix in $\mathcal S$ with $\mathrm{Recov}\ge\rho\,\mathrm{Recov}_{\rm full}$
\State If none qualifies: choose full when in $\mathcal S$; otherwise maximize \mbox{Recov} over $\mathcal S$
\State Train $W^\ast$ under $h$; choose $t^\ast=\arg\max_{t:\mathrm{Env}_t\ge\tau_s}\mathrm{Recov}_t$
\State Choose $\alpha^\ast=\max\{\alpha\in\mathcal A:\mathrm{med}\,\delta_\alpha/\max(\mathrm{med}\,e,10^{-12})\le\tau_d,\ \mathrm{Env}_\alpha\ge\tau_s\}$
\State If checkpoint or scale set is empty, \Return native
\State Run the frozen Original-only closed-loop non-inferiority check at $-10$\pp
\State If passed, \Return $W'_\ell=W_\ell+\alpha^\ast(\kappa/r)B_\ell A_\ell$; otherwise \Return native
\end{algorithmic}
\end{algorithm}

Re-running the frozen controller on the stored metrics with one threshold changed at a time leaves window and scale unchanged
over the ranges of Table~\ref{tab:s-sens}a. Outside them the rule moves as defined: the Faster- and Fast-WAM window widens at
$\rho=0.90$ and $0.95$ (blocks 0--15, then the full backbone) and their scale drops to $\frac12$ at $\tau_d=0.25$. These three
Faster-WAM configurations were trained (same recipe, seed and checkpoint rule) or scaled (from ROOT's checkpoint) and run on 150
further layouts with the default and native (Table~\ref{tab:s-sens}b).

\begin{table}[!htb]
\centering\scriptsize
\setlength{\tabcolsep}{3pt}
\caption{\textbf{The selection is unchanged over a wide range of each threshold, and configurations chosen outside it still
repair.} (a) Threshold values over which the controller's output is unchanged (tested: $\tau_s\in\{0.80,0.85,0.90,0.95,0.975\}$,
$\rho\in\{0.50,0.65,0.80,0.90,0.95\}$, $\tau_d\in\{0.25,0.35,0.50,0.75,1.00\}$; others at 0.90, 0.80, 0.50); UWM and LingBot-VA have
one placement; the main-paper Cosmos Policy adapter uses all blocks (Sec.~\ref{sec:s-probes}). (b) Faster-WAM closed loop, successes of 150; $\Delta$: Recolor+Flat minus the default of the same process, pp
(Tango 95\%); LB: one-sided 95\% lower bound against native, pp (non-inferiority at $-10$).}
\label{tab:s-sens}
\label{tab:s-b2}
\textbf{(a) Range over which the output is unchanged}\\[2pt]
\begin{tabular}{@{}lccccc@{}}
\toprule
Model & window & $\alpha^\ast$ & $\tau_s$ & $\rho$ & $\tau_d$ \\
\midrule
Faster-WAM & 0--7 & 1 & 0.80--0.95 & 0.50--0.80 & 0.35--1.00 \\
Fast-WAM & 0--7 & 1 & 0.80--0.975 & 0.50--0.80 & 0.35--1.00 \\
Cosmos Policy & 0--13 & $\frac12$ & 0.80--0.975 & 0.50--0.95 & 0.50--1.00 \\
UWM & -- & 1 & 0.80--0.975 & -- & 0.25--1.00 \\
LingBot-VA & -- & 1 & 0.80--0.95 & -- & 0.25--1.00 \\
\bottomrule
\end{tabular}

\medskip
\textbf{(b) Closed loop of perturbed configurations}\\[2pt]
\begin{tabular}{@{}llccccc@{}}
\toprule
Threshold & configuration & Orig. & Red. & R+F & $\Delta$ R+F [95\% CI] & LB Orig. / Red. \\
\midrule
--- & native & 145 & 138 & 14 & --- & --- \\
default & 0--7, $\alpha{=}1$ & 145 & 148 & 148 & --- & $-2.8$ / $+3.3$ \\
$\rho=0.90$ & 0--15, $\alpha{=}1$ & 146 & 147 & 146 & $-1.3$ [$-5.1$, $+1.9$] & $-1.7$ / $+1.9$ \\
$\rho=0.95$ & 0--28, $\alpha{=}1$ & 144 & 145 & 145 & $-2.0$ [$-6.1$, $+1.3$] & $-3.4$ / $+1.2$ \\
\midrule
--- & native & 145 & 138 & 15 & --- & --- \\
default & 0--7, $\alpha{=}1$ & 146 & 145 & 148 & --- & $-1.1$ / $+1.2$ \\
$\tau_d=0.25$ & 0--7, $\alpha{=}\frac12$ & 144 & 147 & 135 & $-8.7$ [$-14.3$, $-5.1$] & $-3.4$ / $+3.5$ \\
\bottomrule
\end{tabular}
\end{table}

\section{Additional Results}
\label{sec:s-results}

Closed-loop numbers are successes on paired fresh layouts (150 per model; 100 for Fast-WAM and Cosmos Policy) unless a table says otherwise; R+F is Recolor+Flat and $^\star$ marks a privileged or zero-training reference. Tests: one-sided exact sign tests, Holm-corrected in the registered family; non-inferiority (NI) at $-10$\pp\ (LB: one-sided 95\% lower bound, computed as stated in Sec.~\ref{sec:s-stat}).

\subsection{Main results}

Table~\ref{tab:s-main-full} lists the successes behind main-paper Table~\xr{tab:main}: native, the adapters retrained with the frozen recipe (seed 1 is the main-paper adapter, seeds 2 and 3 retrain it), second inference seeds and the reference arms, plus DiT4DiT (native kept: no window keeps original-scene drift inside the noise envelope) and GR00T N1.7 (no loss, no repair). ROOT raises Recolor+Flat on all six deployed models ($+6.0$ to $+86.7$\pp) with Original and Red-object non-inferior at $-10$\pp. The gain replicates in all three training seeds on Faster-WAM, Fast-WAM, Cosmos Policy and UWM, and every LingBot-VA seed reaches 143--144.

\begin{table}[!htb]
\centering\scriptsize
\setlength{\tabcolsep}{1.6pt}
\caption{\textbf{Successes of every closed-loop run behind the main-paper table.} Models with $n$ layouts in brackets; native is paired across runs; seed 1 is the main-paper adapter; ``---'': not run; L: layouts (Cosmos Policy; official inference seed unless stated); $^\star$privileged or zero-training reference. DiT4DiT: not deployed, no window keeps original-scene drift inside the envelope. Light-WAM Red-object has its own test family. Deployed windows and scales: Faster-WAM blocks 0--28 (0--7: 141--144 on Recolor+Flat), Fast-WAM 0--29 (0--7: 92--95), LingBot-VA all blocks on video tokens, Cosmos Policy all blocks on image tokens at $\alpha=\frac12$ (0--13: 98), Light-WAM 0--16 at $\alpha=\frac12$, UWM its image encoder.}
\label{tab:s-main-full}\label{tab:s-seeds-train}\label{tab:s-main}\label{tab:s-uwm}\label{tab:s-cosmos}\label{tab:s-xmodel}
\begin{minipage}[t]{0.325\linewidth}\vspace{0pt}
\begin{tabular}{@{}l@{\hspace{3pt}}ccccc@{}}
\toprule
Run & R+F & Rec. & Flat & Orig. & Red. \\
\midrule
\multicolumn{6}{@{}l}{\emph{Faster-WAM (150)}} \\
native & 14 & 76 & 119 & 147 & 139 \\
chroma correction$^\star$ & 78 & 141 & 122 & --- & --- \\
late-value restor.$^\star$ & 128 & 141 & 144 & --- & --- \\
ROOT, seed 1 (main) & 144 & 148 & 147 & 146 & 144 \\
ROOT, seed 2 & 147 & 146 & 146 & 144 & 144 \\
ROOT, seed 3 & 148 & 145 & 148 & 147 & 146 \\
chroma correction + ROOT & 144 & 144 & 143 & 146 & 143 \\
native, 2nd inf. seed & 26 & 94 & 128 & 147 & --- \\
ROOT, 2nd inf. seed & 143 & 145 & 146 & 146 & --- \\
\midrule
\multicolumn{6}{@{}l}{\emph{DiT4DiT (150)}} \\
native & 0 & 0 & 123 & 145 & 130 \\
ROOT & \multicolumn{5}{c}{\emph{not deployed (screen)}} \\
\midrule
\multicolumn{6}{@{}l}{\emph{GR00T N1.7 (150)}} \\
native & 132 & 132 & 133 & 128 & 132 \\
ROOT & \multicolumn{5}{c}{\emph{no loss, no repair}} \\
\bottomrule
\end{tabular}
\end{minipage}\hfill
\begin{minipage}[t]{0.325\linewidth}\vspace{0pt}
\begin{tabular}{@{}l@{\hspace{3pt}}ccccc@{}}
\toprule
Run & R+F & Rec. & Flat & Orig. & Red. \\
\midrule
\multicolumn{6}{@{}l}{\emph{Fast-WAM (100)}} \\
native & 16 & 65 & 72 & 94 & 75 \\
chroma correction$^\star$ & 48 & --- & --- & --- & --- \\
ROOT, seed 1 (main) & 88 & 93 & 94 & 93 & 77 \\
ROOT, seed 2 & 90 & 95 & 88 & 92 & 79 \\
ROOT, seed 3 & 89 & 91 & 95 & 93 & 78 \\
native, 2nd inf. seed & 33 & 71 & 84 & 92 & 81 \\
ROOT, 2nd inf. seed & 96 & 93 & 95 & 94 & 84 \\
\midrule
\multicolumn{6}{@{}l}{\emph{UWM (150)}} \\
native & 25 & 26 & 68 & 111 & 88 \\
chroma correction$^\star$ & 18 & --- & --- & --- & --- \\
ROOT, seed 1 (main) & 96 & 92 & 105 & 106 & 103 \\
ROOT, seed 2 & 91 & 88 & 104 & 108 & 88 \\
ROOT, seed 3 & 93 & 90 & 101 & 108 & 96 \\
\midrule
\multicolumn{6}{@{}l}{\emph{Light-WAM (150)}} \\
native & 0 & 4 & 0 & 140 & 105 \\
ROOT 0--16, $\alpha{=}\frac12$, s1 & 21 & 115 & 87 & 142 & 102 \\
ROOT, seed 2 & 24 & 108 & 80 & 142 & --- \\
ROOT, seed 3 & 15 & 110 & 79 & 142 & --- \\
\bottomrule
\end{tabular}
\end{minipage}\hfill
\begin{minipage}[t]{0.325\linewidth}\vspace{0pt}
\begin{tabular}{@{}l@{\hspace{3pt}}ccccc@{}}
\toprule
Run & R+F & Rec. & Flat & Orig. & Red. \\
\midrule
\multicolumn{6}{@{}l}{\emph{Cosmos Policy}} \\
\multicolumn{6}{@{}l}{\hspace{4pt}\emph{layouts 1--10 (/100)}} \\
native & 62 & 94 & 100 & 95 & 94 \\
chroma correction$^\star$ & 91 & --- & --- & --- & --- \\
ROOT, seed 1 (main) & 99 & 98 & 96 & 98 & 94 \\
ROOT, seed 2 & 97 & 99 & 96 & 99 & 94 \\
ROOT, seed 3 & 99 & 100 & 96 & 99 & 94 \\
\multicolumn{6}{@{}l}{\hspace{4pt}\emph{layouts 11--15 (/50)}} \\
native & 39 & 48 & 50 & 45 & 46 \\
chroma correction$^\star$ & 46 & --- & --- & --- & --- \\
ROOT ($\alpha=\frac12$) & 47 & 49 & 50 & 47 & 47 \\
ROOT ($\alpha=1$) & 48 & --- & --- & 47 & 46 \\
\multicolumn{6}{@{}l}{\hspace{4pt}\emph{layouts 1--10, 3rd inf. seed (/100)}} \\
native & 68 & --- & 99 & 99 & --- \\
ROOT & 97 & --- & 99 & 98 & --- \\
\midrule
\multicolumn{6}{@{}l}{\emph{LingBot-VA (150)}} \\
native & 135 & 144 & 145 & 149 & 145 \\
ROOT, seed 1 (main) & 144 & 135 & 140 & 144 & 146 \\
ROOT, seed 2 & 143 & 139 & 142 & 144 & 144 \\
ROOT, seed 3 & 143 & 140 & 144 & 144 & 147 \\
\bottomrule
\end{tabular}
\end{minipage}
\end{table}

\paragraph{Tests.} On Faster-WAM the second inference seed gives 143 against 26 on Recolor+Flat (118/1 discordant, $p=1.8\times10^{-34}$), with Original non-inferior (LB $-3.3$\pp). Table~\ref{tab:s-tests} lists the primary tests.

\begin{table}[!htb]
\centering\footnotesize
\setlength{\tabcolsep}{3pt}
\caption{\textbf{Tests of ROOT against native.} (a) All five conditions for the single-seed Faster-WAM and UWM runs: discordant = ROOT-only/native-only successes; difference with 95\% interval (Faster-WAM as proportion, UWM in pp; LB: lower bound of the one-sided 95\% non-inferiority test, NI); $p$: one-sided exact sign test, unadjusted. (b) Recolor+Flat for the other runs, with the safety checks: lower bound of the paired difference to native (pp) or NI where only the verdict is available; Holm: Holm-adjusted within the registered family.}
\label{tab:s-tests}
\par\noindent\emph{(a) Faster-WAM and UWM, five conditions}\par\smallskip
{\scriptsize
\begin{tabular}{@{}lccccc@{}}
\toprule
 & R+F & Recolor & Flat & Original & Red-object \\
\midrule
Faster-WAM: discordant & 131/1 & 73/1 & 30/2 & 2/3 & 6/1 \\
Faster-WAM: difference [95\% CI] & $+0.867$ [0.807, 0.920] & $+0.480$ [0.400, 0.560] & $+0.187$ [0.120, 0.253] & $-0.007$ (LB $-0.033$) NI & $+0.033$ (LB $+0.007$) NI \\
Faster-WAM: $p$ & $2.4\times10^{-38}$ & $4.0\times10^{-21}$ & $1.2\times10^{-7}$ & & \\
\midrule
UWM: discordant & 77/6 & 74/8 & 44/7 & 7/12 & 26/11 \\
UWM: difference, pp [95\% CI] & $+47.3$ [38, 56] & $+44.0$ [34.7, 53.3] & $+24.7$ [16.7, 33.3] & $-3.3$ (LB $-8.0$) & $+10.0$ (LB $+3.3$) \\
\bottomrule
\end{tabular}}
\par\medskip\noindent\emph{(b) Other runs, Recolor+Flat}\par\smallskip
\begin{tabular}{@{}llcccll@{}}
\toprule
Model & Run & Diff [95\% CI] & Discordant & $p$ & Original & Red-object \\
\midrule
Faster-WAM & 2nd inference seed & $+78$ [70.7, 84.7] & 118/1 & $1.8\times10^{-34}$ & LB $-3.3$ & --- \\
Fast-WAM & seed 1 & $+72$ & 74/2 & $3.9\times10^{-20}$ & LB $-5$ & LB $-2$ \\
 & 2nd inference seed & $+63$ [53, 73] & --- & Holm & NI & NI \\
Cosmos Policy & layouts 1--10 & $+37$ [27, 47] & 38/1 & Holm & NI & NI \\
 & layouts 11--15 & $+16$ [4, 30] & 10/2 & 0.019 & NI & NI \\
 & layouts 1--10, 3rd inference seed & $+29$ [19, 39] & --- & Holm & NI & --- \\
LingBot-VA & seed 1 & $+6.0$ [0.7, 12.0] & 14/5 & 0.032 (Holm) & LB $-6.0$ & --- \\
Light-WAM & seed 1 & $+14.0$ [8.7, 20.0] & --- & $<10^{-5}$ (Holm) & NI & NI ($p=0.022$) \\
\bottomrule
\end{tabular}
\end{table}

\paragraph{Cross-model summary.} Averaged over the six deployed models, ROOT adds 43.8\pp\ on Recolor+Flat ($+86.7$, $+72.0$, $+47.3$, $+37.0$, $+6.0$, $+14.0$\pp\ on Faster-WAM, Fast-WAM, UWM, Cosmos Policy, LingBot-VA and Light-WAM) and moves Original by $-0.7$, $-1.0$, $-3.3$, $+3.0$, $-3.3$ and $+1.3$\pp\ in the same order, at most 3.3\pp\ in either direction.

\subsection{Baselines and ablations}

\paragraph{Same-budget baselines.} Aug-BC (same renderings, adapter placement, rank, steps and optimizer; only the loss differs) and two re-implemented published objectives, an input-perturbation objective in the style of RobustVLA and representation anchoring to the base model, are compared with ROOT in Table~\ref{tab:s-augbc} (native and ROOT counts: Table~\ref{tab:s-main-full}). ROOT is superior on Recolor+Flat, Recolor and Flat in every matched run (anchoring, seed 3, Flat: not significant after Holm), no baseline is non-inferior to native on Red-object in any seed, and under ROOT's scale rule every baseline deploys only at $\alpha=\frac18$ (in ROOT's own order none does: no baseline checkpoint stays inside the envelope at full strength, Figure~\ref{fig:s-screen}) and ROOT is better by 63--70\pp\ on Recolor+Flat (18 of 18 tests). Preprocessing and steering, whose Recolor+Flat results are in main-paper Table~\xr{tab:alt}: CLAHE and gray-world leave Faster-WAM at 11 and 17 of 150 (native 14); on Cosmos Policy they give 108 and 131 against 101 (ROOT 146); late-site steering ($\gamma=\frac12$) reaches 21 of 150 on Recolor+Flat, 81 on Recolor and 125 on Flat, with Original 148 against 147.

\begin{table}[!htb]
\centering\footnotesize
\setlength{\tabcolsep}{2.8pt}
\caption{\textbf{ROOT beats same-budget action-label training and two published objectives on all three shifted conditions, and passes the Red-object check that they fail.} (a) Single training seed. $^\times$fails non-inferiority to native; $^r$our re-implementation (input branch of an input-perturbation objective in the style of RobustVLA, without adversarial PGD and output branch; representation anchoring with the published weight, not rescaled to the Wan backbone); ROOT vs.\ each baseline is a separate family (one-sided exact sign test, Holm over the three conditions at 0.05 for Aug-BC and at 0.025 for each re-implemented objective); safety by the Tango score test at $-10$\pp. (b) Re-implementations across training seeds; Fast-WAM rows are successes of 150 layouts (first 15 per task); NI: paired Tango non-inferiority to native; $^d$Flat not significant after Holm. (c) Arms configured by ROOT's scale rule (Faster-WAM, $\alpha=\frac18$, of 150; scale-then-select: checkpoint chosen after scaling; own rule: the arm's own checkpoint rule); last column: ROOT$-$arm on R+F with Tango 95\% CI.}
\label{tab:s-augbc}\label{tab:s-pubbase-seeds}\label{tab:s-ctrl-arms}
\par\noindent\emph{(a) Single seed}\par\smallskip
\begin{tabular}{@{}lccccc@{}}
\toprule
Arm & Recolor+Flat & Recolor & Flat & Original & Red-object \\
\midrule
\multicolumn{6}{@{}l}{\emph{Faster-WAM (/150)}} \\
Aug-BC & 128 & 141 & 126 & 139 & 122$^\times$ \\
ROOT $-$ Aug-BC, pp [95\% CI] & $+10.7$ [4.7, 16.7] & $+4.7$ [0.7, 9.3] & $+14.0$ [8.7, 20.0] & & \\
discordant; $p$ (Holm) & 20/4; 0.0015 & 9/2; 0.033 & 22/1; $9\times10^{-6}$ & & \\
\midrule
\multicolumn{6}{@{}l}{\emph{Fast-WAM (/100)}} \\
Aug-BC & 73 & 81 & 72 & 78$^\times$ & 69$^\times$ \\
ROOT $-$ Aug-BC, pp [95\% CI] & $+15$ [5, 25] & $+12$ [3, 21] & $+22$ [14, 30] & & \\
discordant; $p$ (Holm) & 22/7; 0.008 & 17/5; 0.008 & 22/0; $<0.001$ & & \\
\midrule
\multicolumn{6}{@{}l}{\emph{Re-implementations (Faster-WAM, /150; ROOT and native as above)}} \\
RobustVLA-style input perturbation$^r$ & 130 & 140 & 136 & 137$^\times$ & 127$^\times$ \\
ROOT $-$ it, pp [95\% CI] & $+9.3$ [3.3, 15.3] & $+5.3$ [1.3, 10.0] & $+7.3$ [2.0, 12.7] & & \\
discordant; $p$ (Holm) & 18/4; 0.0065 & 10/2; 0.019 & 14/3; 0.013 & & \\
anchor regularization$^r$ & 114 & 133 & 126 & 131$^\times$ & 115$^\times$ \\
ROOT $-$ it, pp [95\% CI] & $+20.0$ [12.7, 27.3] & $+10.0$ [4.7, 15.3] & $+14.0$ [8.0, 20.0] & & \\
discordant; $p$ (Holm) & 33/3; $<10^{-5}$ & 17/2; 0.00036 & 23/2; $2\times10^{-5}$ & & \\
\bottomrule
\end{tabular}
\par\medskip\noindent\emph{(b) Re-implementations across seeds}\par\smallskip
{\scriptsize
\begin{tabular}{@{}lllcccccccl@{}}
\toprule
Model & Arm & Seed & R+F & R & Flat & Orig. & Red-obj. & NI Orig. & NI Red-obj. & Verdict \\
\midrule
Faster-WAM & Input-pert. & 1 & 130 & 140 & 136 & 137 & 127 & no & no & ROOT superior on 3/3 \\
 &  & 2 & 112 & 127 & 107 & 131 & 112 & no & no & ROOT superior on 3/3 \\
 &  & 3 & 108 & 126 & 130 & 140 & 121 & yes & no & ROOT superior on 3/3 \\
\midrule
Faster-WAM & Rep.-anchor. & 1 & 114 & 133 & 126 & 131 & 115 & no & no & ROOT superior on 3/3 \\
 &  & 2 & 116 & 129 & 120 & 129 & 113 & no & no & ROOT superior on 3/3 \\
 &  & 3 & 130 & 128 & 139 & 141 & 112 & yes & no & ROOT superior on 2/3$^d$ \\
\midrule
Fast-WAM & Input-pert. & 1 & 101 & 119 & 105 & 114 & 91 & no & no & ROOT superior on 3/3 \\
 &  & 2 & 97 & 112 & 101 & 114 & 90 & no & no & ROOT superior on 3/3 \\
\midrule
Fast-WAM & Rep.-anchor. & 1 & 112 & 114 & 106 & 122 & 99 & no & no & ROOT superior on 3/3 \\
 &  & 2 & 89 & 111 & 105 & 115 & 92 & no & no & ROOT superior on 3/3 \\
\bottomrule
\end{tabular}}
\par\medskip\noindent\emph{(c) Under ROOT's controller}\par\smallskip
\begin{tabular}{@{}llcccccc@{}}
\toprule
Arm & Checkpoint & Orig. & Flat & Rec. & R+F & Red. & ROOT$-$arm, R+F (\pp) \\
\midrule
Native & -- & 147 & 119 & 76 & 14 & 139 & -- \\
ROOT & -- & 146 & 147 & 148 & 144 & 144 & -- \\
\midrule
Aug-BC & scale-then-select & 147 & 130 & 108 & 39 & 140 & $+70.0$ [61.8, 76.9] \\
 & own rule & 145 & 132 & 114 & 48 & 142 & $+64.0$ [55.7, 71.4] \\
Input-pert.$^r$ & scale-then-select & 145 & 129 & 112 & 49 & 137 & $+63.3$ [55.0, 70.8] \\
 & own rule & 147 & 134 & 115 & 46 & 138 & $+65.3$ [57.0, 72.6] \\
Rep.-anchor.$^r$ & scale-then-select & 146 & 126 & 94 & 41 & 141 & $+68.7$ [60.4, 75.7] \\
 & own rule & 145 & 130 & 93 & 39 & 140 & $+70.0$ [61.8, 76.9] \\
\bottomrule
\end{tabular}
\end{table}

\paragraph{Ablations.} Table~\ref{tab:s-abl-full}a extends main-paper Table~\xr{tab:ablate} with lower bounds and checkpoint statistics; each variant is paired with the ROOT run of its own process (Stage 2: 148 of 150; Stage 1, the scale variant: 147 of 150). Dropping the clean-view term leaves one of 16 checkpoints inside the envelope (0.91; ROOT 1.00), and adding the demonstration-action loss leaves none (best 0.64). Two further variants (panel b): a 2D per-channel gain on the table region of the original frame (no renderer) ties ROOT (joint over two seeds $+1.4$\pp\ [$-1.7$, $+4.5$] on LIBERO-Plus Background), the image-space recoloring used on the real robot (Sec.~S7); training on the model's own original-scene rollouts, with no demonstrations, ties ROOT and is the variant used for GE-Act (Sec.~S6).

\begin{table}[!htb]
\centering\footnotesize
\setlength{\tabcolsep}{2.5pt}
\caption{\textbf{Removing the clean-view term, using blocks 16--28 or rank 8 costs 4.0--13.3\pp\ on Recolor+Flat, and one eighth of the update costs 67.3\pp; Original and Red-object stay non-inferior.} (a) Closed-loop and offline ablations on Faster-WAM (successes of 150). $\Delta$: R+F minus the ROOT re-run of the same process, pp (Tango 95\%); LB: one-sided 95\% lower bound of Original or Red-object against native, pp (NI at $-10$). Offline: saved checkpoints satisfying the envelope constraint of 16, selected step, held-out recovery and envelope fraction (without an eligible checkpoint: best over the curve). The scale variant is evaluated from ROOT's checkpoint at one eighth of the update. (b) Image-space and rollout variants: LIBERO-Plus Background (/289; two-sided exact sign test, tie = NI at $L>-8$\pp) and fresh-layout Recolor+Flat (/150; tie at $L>-10$\pp); Original vs.\ native 147 (LB).}
\label{tab:s-abl-full}\label{tab:s-ablations}
\par\noindent\emph{(a) Closed-loop ablations}\par\smallskip
{\setlength{\tabcolsep}{2pt}
\begin{tabular}{@{}l ccc c cc ccc@{}}
\toprule
& Orig. & Red. & R+F & $\Delta$ R+F [95\% CI] & LB Orig. & LB Red. & ckpts & step & recov./env. \\
\midrule
\multicolumn{10}{@{}l}{\emph{Stage 2}} \\
native & 145 & 138 & 16 & --- & --- & --- & --- & --- & --- \\
ROOT (0--7, rank 32, $\alpha{=}1$) & 145 & 146 & 148 & --- & $-2.8$ & $+1.7$ & 16 & 19,500 & 0.543 / 1.000 \\
no clean-view term & 144 & 145 & 138 & $-6.7$ [$-11.8$, $-3.7$] & $-4.1$ & $+1.2$ & 1 & 1,500 & 0.401 / 0.907 \\
+ demonstration-action loss & \multicolumn{3}{c}{not run in closed loop} & --- & --- & --- & 0 & --- & best 0.402 / 0.640 \\
blocks 16--28 & 147 & 140 & 128 & $-13.3$ [$-19.9$, $-8.2$] & $-1.2$ & $-2.4$ & 13 & 18,000 & 0.376 / 0.947 \\
rank 8 & 143 & 148 & 142 & $-4.0$ [$-8.8$, $-0.4$] & $-4.3$ & $+3.3$ & 16 & 21,000 & 0.522 / 0.993 \\
\midrule
\multicolumn{10}{@{}l}{\emph{Stage 1}} \\
native & 145 & 137 & 14 & --- & --- & --- & --- & --- & --- \\
ROOT & 145 & 145 & 147 & --- & $-2.3$ & $+1.4$ & 16 & 19,500 & 0.543 / 1.000 \\
scale $\frac18$ & 146 & 144 & 46 & $-67.3$ [$-74.5$, $-59.0$] & $-1.7$ & $+2.5$ & \multicolumn{3}{c}{ROOT's checkpoint} \\
\bottomrule
\end{tabular}}
\par\medskip\noindent\emph{(b) Image-space and rollout variants}\par\smallskip
\begin{tabular}{@{}l cc c c cc c@{}}
\toprule
& \multicolumn{4}{c}{LIBERO-Plus Background (/289)} & \multicolumn{2}{c}{Recolor+Flat (/150)} & Original \\
\cmidrule(lr){2-5}\cmidrule(lr){6-7}
Variant & variant & ROOT & $\Delta$ [95\% CI], pp & $p$ & variant & ROOT & (LB) \\
\midrule
2D gain in table mask, seed 1 & 182 & 178 & $+1.4$ [$-2.8$, $+5.9$] & 0.64 & 143 & 145 & 141 ($-6.7$) \\
2D gain, seed 2 (vs.\ ROOT seed 2) & 183 & 179 & $+1.4$ [$-2.4$, $+5.2$] & 0.60 & 143 & 147 & 146 ($-2.7$) \\
2D gain, joint over two seeds & --- & --- & $+1.4$ [$-1.7$, $+4.5$] & 0.14 & \multicolumn{2}{c}{$-2.0$ [$-4.7$, $+0.7$]} & --- \\
\midrule
own rollouts, no demonstrations & 183 & 177 & $+2.1$ [$-1.4$, $+5.5$] & 0.34 & 144 & 142 & 143 ($-5.3$) \\
\bottomrule
\end{tabular}
\end{table}

\paragraph{Texture views.} Adding procedural high-frequency texture views (disjoint from all test textures) to the ROOT pairs raises success on the held-out texture cells of the appearance sweep from 100 to 143 of 150 (Holm), on LIBERO-Plus table textures from 101 to 124 of 168 and on LIBERO-Plus Background from 79 to 109 of 121 (a second training seed: 144, 134 and 111), with Original non-inferior and Recolor+Flat at 139--144.

\subsection{Safety and behavior}
\label{sec:s-screen}

\begin{figure}[!htb]
\centering
\includegraphics[width=\linewidth]{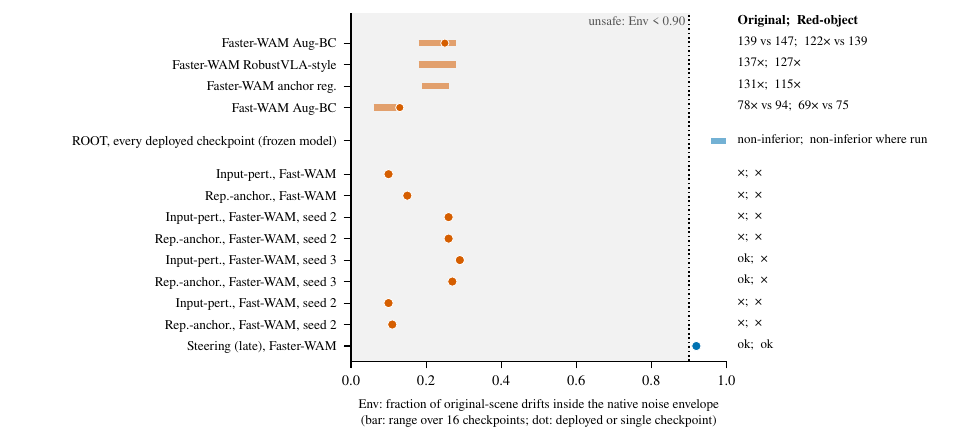}
\caption{\textbf{An offline envelope fraction below 0.90 predicts a closed-loop safety failure.} Env: fraction of held-out original-scene drifts inside the native noise envelope. Upper four arms: range over the 16 saved checkpoints (dot: deployed checkpoint where read); lower arms: further arms and seeds with Env read from offline records written before their closed loop. Right: non-inferiority to native at $-10$\pp\ (successes in the upper block, ok / $\times$ in the lower block; $\times$ fails). ROOT arms are excluded from the lower block (Env selected them); all nine lower-block arms are classified correctly.}
\label{fig:s-screen}
\end{figure}

\paragraph{Offline safety screen.} The envelope fraction Env (main paper, Sec.~\xr{sec:method}) needs no shifted scene. Read on the label-supervised repairs, whose checkpoints were not selected with it, it separates the repairs we compared (Figure~\ref{fig:s-screen}): no label-supervised arm kept more than 28\% of original-scene drifts inside the envelope at any of its 16 checkpoints and all four failed at least one closed-loop original-scene safety test, whereas every deployed ROOT kept 95--100\% and passed.

\subsection{Placement and controller runs}
\label{sec:s-probes}

\begin{figure}[!htb]
\centering
\includegraphics[width=0.86\linewidth]{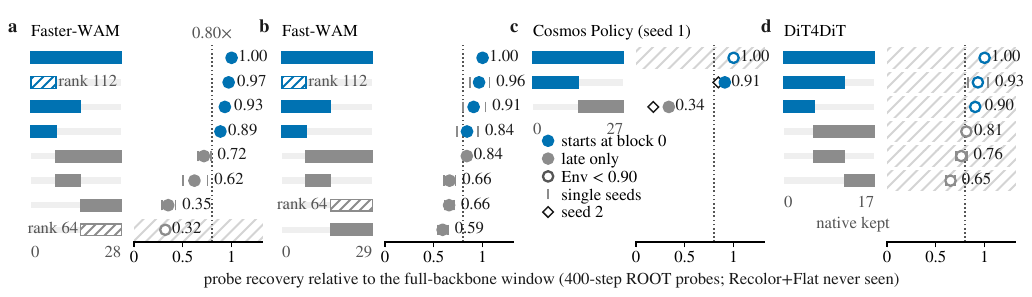}%
\caption{\textbf{Input-side windows repair; late-only windows do not.} 400-step placement probes on held-out random appearances (never Recolor+Flat): recovery of each window relative to the full-backbone window. Left bars: the window's span on the model's block axis (blue: starts at block 0; gray: does not; hatched: rank-matched capacity controls). Filled dots: inside the original-scene safety envelope (Env $\ge0.90$); open dots: outside it. Ticks: single seeds; diamond: Cosmos seed 2; dotted: the 0.80 rule. (d) DiT4DiT (two-seed means): no window passes the screen (best Env 0.87), so the native model is kept.}
\label{fig:s-probes}
\end{figure}

Table~\ref{tab:s-windows} lists the probe values. Across the eight arms of Faster- and Fast-WAM, parameter count correlates only 0.46 and 0.47 with Recov (Spearman), so placement, not capacity, drives recovery. On Cosmos Policy blocks 0--13 are inside the envelope with both seeds, and the input-side window recovers 0.91 and 0.85 of the full window against 0.34 and 0.18 for the late-only one. On DiT4DiT, too, windows that contain block 0 rank highest (Spearman 0.94).

\begin{table}[!htb]
\centering\footnotesize
\setlength{\tabcolsep}{3pt}
\caption{\textbf{Window probes.} 400 steps, held-out random appearances, never Recolor+Flat. Recov: median first-chunk recovery; Env: fraction of original-scene drifts inside the noise envelope (unsafe if $<0.90$, $^u$). (a, b) Faster-WAM, Fast-WAM and DiT4DiT: mean of two training seeds; Cosmos Policy: recipe at scale $\frac12$, seed 1 / seed 2; late windows: 16--28, 16--29, 14--27. Main-paper ratios use Cosmos Policy seed 1. (c) Light-WAM: mean of two training seeds; Recov and Env at $\alpha=1$ for the primary flow and at $\alpha=\frac12$ for the fallback; $^s$selectable arm.}
\label{tab:s-windows}\label{tab:s-lw}
\begin{minipage}[t]{0.555\linewidth}\vspace{0pt}
\emph{(a) Faster-WAM, Fast-WAM, Cosmos Policy}\par\smallskip
\begin{tabular}{@{}l c cc cc@{}}
\toprule
& & \multicolumn{2}{c}{Faster-WAM} & \multicolumn{2}{c}{Fast-WAM} \\
\cmidrule(lr){3-4}\cmidrule(lr){5-6}
Window & rank & \mbox{Recov} & Env & \mbox{Recov} & Env \\
\midrule
full (0--28 / 0--29) & 32 & \textbf{0.313} & 0.947 & \textbf{0.276} & 0.987 \\
0--7 & 112 & 0.304 & 0.943 & 0.266 & 0.983 \\
0--15 & 32 & 0.292 & 0.970 & 0.251 & 0.997 \\
0--7 & 32 & 0.278 & 0.973 & 0.232 & 0.993 \\
8--end & 32 & 0.225 & 0.910 & 0.232 & 0.983 \\
8--15 & 32 & 0.194 & 0.947 & 0.183 & 0.993 \\
late & 32 & 0.110 & 0.913 & 0.163 & 0.990 \\
late & 64 & 0.100 & 0.873$^u$ & 0.181 & 0.987 \\
\midrule
\multicolumn{2}{@{}l}{\emph{Cosmos Policy (seed 1 / seed 2)}} & \mbox{Recov} & Env & \mbox{Recov} & Env \\
0--27 & 32 & 0.111 & 0.780$^u$ & 0.159 & 0.987 \\
0--13 & 32 & 0.101 & 0.987 & 0.135 & 0.993 \\
14--27 & 32 & 0.037 & 0.973 & 0.028 & 1.000 \\
\bottomrule
\end{tabular}
\end{minipage}\hfill
\begin{minipage}[t]{0.425\linewidth}\vspace{0pt}\scriptsize
\emph{(b) DiT4DiT}\par\smallskip
\begin{tabular}{@{}lcccc@{}}
\toprule
\multicolumn{5}{@{}l}{\emph{DiT4DiT}} \\
Window & rank & \mbox{Recov} & Env & \mbox{Recov}/full \\
\midrule
0--17 (full) & 32 & 0.275 & 0.670$^u$ & 1.00 \\
0--11 & 32 & 0.256 & 0.600$^u$ & 0.93 \\
0--5 & 32 & 0.248 & 0.563$^u$ & 0.90 \\
6--17 & 32 & 0.223 & 0.763$^u$ & 0.81 \\
6--11 & 32 & 0.209 & 0.873$^u$ & 0.76 \\
12--17 & 32 & 0.179 & 0.723$^u$ & 0.65 \\
\bottomrule
\end{tabular}

\medskip\emph{(c) Light-WAM}\par\smallskip
\begin{tabular}{@{}l c c c c@{}}
\toprule
Window & \mbox{Recov} ($\alpha{=}1$) & Env ($\alpha{=}1$) & \mbox{Recov} ($\alpha{=}\frac12$) & Env ($\alpha{=}\frac12$) \\
\midrule
0--8$^s$ & 0.268 & 0.737 & 0.196 & 0.993 \\
0--16$^s$ & 0.322 & 0.787 & 0.252 & 0.990 \\
0--24$^s$ (full) & 0.396 & 0.697 & 0.296 & 0.963 \\
9--16 & 0.253 & 0.843 & 0.173 & 0.990 \\
17--24 & 0.253 & 0.743 & 0.197 & 0.983 \\
9--24 & 0.312 & 0.707 & 0.232 & 0.977 \\
\bottomrule
\end{tabular}
\end{minipage}
\end{table}

\paragraph{Light-WAM.}\label{sec:s-lw}
Light-WAM did not shape the procedure or its thresholds. Its action head is deterministic, so the envelope $e_i$ is the largest first-chunk distance under four draws of $\pm2$-pixel jitter of the original frame. At full strength no window reaches Env $\ge0.90$; at $\alpha=\frac12$ every arm is safe (Env 0.96--0.99) and the window rule keeps blocks 0--16 (Recov 0.252, 0.85 of the full window), with scale $\frac12$ (drift ratio 0.30). Closed loop (Tables~\ref{tab:s-seeds-train}, \ref{tab:s-tests}): Recolor 4$\to$115 ($+74.0$\pp\ [66.7, 80.7], Holm $p<10^{-6}$), Flat 0$\to$87 ($+58.0$\pp\ [50.0, 66.0]), Original 140$\to$142 ($+1.3$\pp\ [$-2.7$, 5.3]; NI, Holm $p<10^{-4}$); over three seeds Recolor+Flat is 21, 24 and 15 and Recolor 115, 108 and 110. Red-object (own test family) is 105 against 102 ($-2.0$\pp\ [$-9.3$, $+5.3$]; NI $p=0.022$); recomputing the red-instruction embeddings with Light-WAM's own pipeline leaves behavior unchanged (141 against 140 of 150).

\paragraph{Controller runs.}\label{sec:s-prosp}
In six prospective runs (Faster-WAM and Streaming-WAM, three training seeds each) one frozen controller took every decision (thresholds $\tau_s=0.90$, $\rho=0.80$, $\tau_d=0.50$; scale grid $\{1,\frac12,\frac14,\frac18\}$; five prefix windows; rank 32; 23{,}400 steps), with no fallback branch, on disjoint probe, checkpoint-selection, scale and rollout splits; the rollouts use 150 layouts no earlier selection or evaluation touched. Streaming-WAM is a public checkpoint of the Fast-WAM lineage (Wan2.2-5B); natively it succeeds in 28 of 30 original-scene episodes and 9 of 30 on Recolor+Flat. Original and Red-object non-inferiority hold in all six runs (Tango $z$ 3.18--3.95 and 3.72--4.68); native Streaming-WAM already scores 137--138 of 150 on Flat, so that test does not reject (Table~\ref{tab:s-prosp}a).

\paragraph{Re-scoring on independent layouts.}\label{sec:s-fresh}
We drew 150 further layouts from the same sampler (at least 2\,mm from every earlier layout, no model queried) and re-scored the deployed adapters and the native models in the original scene, under Recolor+Flat and under Red-object: all six superiority tests (Holm over six) and all twelve safety tests (Holm over twelve) reject (Table~\ref{tab:s-prosp}b), and using the second native replicate changes no verdict. Pooled over seeds, Recolor+Flat success is 8.7\%$\to$96.4\% on Faster-WAM and 27.3\%$\to$83.1\% on Streaming-WAM, and the Original-scene difference is $+0.9$\pp\ ($-1.6$, $+3.6$) and $+0.4$\pp\ ($-2.9$, $+3.8$) (layout-bootstrap 95\% intervals).

\begin{table}[!htb]
\centering\footnotesize
\setlength{\tabcolsep}{2.5pt}
\caption{\textbf{Prospective controller runs on the shared layouts (a) and on 150 independent layouts (b).} Successes out of 150, Native$\to$ROOT as deployed; Seed = training seed. Sign tests for Recolor+Flat, Recolor and Flat: ROOT-only/Native-only wins among discordant layouts, one-sided exact sign test, Holm over the three; $\checkmark$ rejected at 0.05, $\times$ not rejected. (b) Recolor+Flat test: Holm $p$ over six runs; Original and Red-object non-inferiority at $-10$\pp, Holm over twelve tests, all rejected ($p\le0.0022$); mean rows average the three seeds; the layout counts are ROOT-better/Native-better layouts.}
\label{tab:s-prosp}\label{tab:s-fresh}
\par\noindent\emph{(a) Shared layouts}\par\smallskip
\begin{tabular}{@{}l c c c c c c c c c c c@{}}
\toprule
Seed & Blocks & $t$ & $\alpha$ & Original & Flat & Recolor & Recolor+Flat & Red-object & R+F test & Recolor test & Flat test \\
\midrule
\multicolumn{12}{@{}l}{\emph{Faster-WAM}} \\
4 & 0--3 & 21k   & 1 & 145$\to$144 & 126$\to$146 & 87$\to$146 & 15$\to$143 & 138$\to$145 & 128/0 $\checkmark$ & 61/2 $\checkmark$ & 22/2 $\checkmark$ \\
5 & 0--3 & 23.4k & 1 & 144$\to$145 & 126$\to$146 & 86$\to$146 & 16$\to$143 & 139$\to$144 & 127/0 $\checkmark$ & 61/1 $\checkmark$ & 22/2 $\checkmark$ \\
6 & 0--3 & 22.5k & 1 & 144$\to$143 & 126$\to$147 & 87$\to$143 & 14$\to$141 & 139$\to$141 & 127/0 $\checkmark$ & 57/1 $\checkmark$ & 22/1 $\checkmark$ \\
\midrule
\multicolumn{12}{@{}l}{\emph{Streaming-WAM}} \\
4 & 0--3 & 23.4k & 0.5 & 146$\to$145 & 137$\to$144 & 101$\to$142 & 35$\to$127 & 128$\to$135 & 92/0 $\checkmark$ & 45/4 $\checkmark$ & 11/4 $\times$ \\
5 & 0--3 & 21k   & 0.5 & 146$\to$145 & 137$\to$144 & 103$\to$137 & 36$\to$128 & 127$\to$132 & 93/1 $\checkmark$ & 41/7 $\checkmark$ & 11/4 $\times$ \\
6 & 0--7 & 23.4k & 0.5 & 145$\to$143 & 138$\to$143 & 101$\to$141 & 36$\to$127 & 129$\to$131 & 92/1 $\checkmark$ & 45/5 $\checkmark$ & 10/5 $\times$ \\
\bottomrule
\end{tabular}
\par\medskip\noindent\emph{(b) Independent layouts}\par\smallskip
\begin{tabular}{@{}llccccc@{}}
\toprule
Model & Seed & Window, $\alpha$ & Original & Recolor+Flat & Red-object & Recolor+Flat test (Holm $p$) \\
\midrule
Faster-WAM & 4 & 0--3, 1 & 143$\to$145 & 13$\to$145 & 138$\to$142 & 133/1 ($2.5\times10^{-38}$) \\
 & 5 & 0--3, 1 & 143$\to$144 & 13$\to$146 & 138$\to$141 & 133/0 ($5.5\times10^{-40}$) \\
 & 6 & 0--3, 1 & 143$\to$144 & 13$\to$143 & 138$\to$145 & 130/0 ($3.7\times10^{-39}$) \\
 & mean & & 143$\to$144.3 & 13$\to$144.7 & 138$\to$142.7 & 135/1 \\
\midrule
Streaming-WAM & 4 & 0--3, $\frac12$ & 143$\to$141 & 41$\to$127 & 131$\to$132 & 86/0 ($3.9\times10^{-26}$) \\
 & 5 & 0--3, $\frac12$ & 143$\to$145 & 41$\to$123 & 131$\to$133 & 83/1 ($4.4\times10^{-24}$) \\
 & 6 & 0--7, $\frac12$ & 143$\to$145 & 41$\to$124 & 131$\to$129 & 83/0 ($2.1\times10^{-25}$) \\
 & mean & & 143$\to$143.7 & 41$\to$124.7 & 131$\to$131.3 & 93/1 \\
\bottomrule
\end{tabular}
\end{table}

\paragraph{Narrow window.} On Faster-WAM blocks 0--7 equal the full adapter on Recolor+Flat (4/4 discordant; 0\pp\ [$-4$, $+4$]; NI $p=0.015$), the pooled confirmatory test on 150 new layouts confirms non-inferiority (Table~\ref{tab:s-narrow}, pooled row), and Red-object is 141 of 150. On Fast-WAM they give 95 against 88 ($+7$\pp\ [$+1$, $+14$]) with both tests rejected in every seed, and the scale rule on that adapter returns 1 (ratio 0.17). On Cosmos Policy blocks 0--13 give 98 against 99 of the full adapter and 98 against 62 of native (37/1), Original 98 against 95; Table~\xr{tab:main} reports the full-window adapter.

\begin{table}[!htb]
\centering\footnotesize
\setlength{\tabcolsep}{3pt}
\caption{\textbf{Narrow window against the full adapter in closed loop.} Blocks 0--7 (Cosmos Policy 0--13); Faster-WAM /150, Fast-WAM and Cosmos Policy /100; single inference seed; native and full arms paired across runs. Seeds 2 and 3 are retrained narrow adapters; NI $p_\mathrm{Holm}$: non-inferiority to the full adapter at $-5$\pp\ on Recolor+Flat (family: that test, Original non-inferiority to native at $-10$\pp, Recolor+Flat superiority to native). $^p$Confirmatory test on 150 new layouts (old layouts excluded) pooled over seeds: mean narrow$-$full $-0.7$\pp, one-sided 95\% lower bound $-3.3$\pp\ against $-5$\pp; every narrow seed keeps Original non-inferior to native; per-seed differences $0.0/-2.7/+0.7$\pp; native 15/150, full adapter 146.}
\label{tab:s-narrow}
\begin{tabular}{@{}llccccll@{}}
\toprule
Model & Arm & R+F & Recolor & Flat & Orig. & NI $p_\mathrm{Holm}$ & Verdict \\
\midrule
Faster-WAM & native & 14 & 76 & 119 & 147 & & \\
 & full (0--28) & 144 & 148 & 147 & 146 & & \\
 & 0--7, seed 1 & \textbf{144} & 146 & 145 & 144 & 0.015 & NI shown \\
 & 0--7, seed 2 & 141 & --- & --- & 146 & 0.091 & in pooled test \\
 & 0--7, seed 3 & 142 & --- & --- & 145 & 0.061 & in pooled test \\
 & 0--7, pooled$^p$ & 146/142/147 & --- & --- & ok & --- & NI shown \\
\midrule
Fast-WAM & native & 16 & 65 & 72 & 94 & & \\
 & full (0--29) & 88 & 93 & 94 & 93 & & \\
 & 0--7, seed 1 & \textbf{95} & 95 & 87 & 94 & 0.002 & NI shown \\
 & 0--7, seed 2 & 92 & --- & --- & 92 & 0.007 & NI shown \\
 & 0--7, seed 3 & 92 & --- & --- & 94 & 0.006 & NI shown \\
\midrule
Cosmos Policy & native & 62 & 94 & 100 & 95 & & \\
 & full (all blocks, $\alpha{=}\frac12$) & 99 & 98 & 96 & 98 & & \\
 & 0--13 ($\alpha{=}1$) & 98 & 95 & 100 & 98 & --- & --- \\
\bottomrule
\end{tabular}
\end{table}

\paragraph{Scale rule.} Recovery and drift rise monotonically with $\alpha$ on all five models, and the rule returns the largest scale inside the admissible region: full strength on four models and $\frac12$ on Cosmos Policy, whose $\alpha=1$ arm was equally safe in closed loop (Original 47, Red-object 46, Recolor+Flat 48 of 50).

\begin{table}[!htb]
\centering\footnotesize
\caption{\textbf{The scale rule on the deployed checkpoints.} Ratio: median original-scene drift / median native noise envelope (pass $\le0.5$); env.: fraction of original-scene drifts inside the envelope (admissible: ratio $\le0.5$, env.\ $\ge0.90$). Faster-WAM, Fast-WAM, UWM, LingBot-VA: rule hold-out unused by any selection; Cosmos Policy: selection hold-out.}
\label{tab:s-dose}
\begin{tabular}{@{}l cccc c c c@{}}
\toprule
& \multicolumn{4}{c}{recovery at $\alpha=1,\frac12,\frac14,\frac18$} & ratio ($\alpha=1$) & env.\ ($\alpha=1$) & rule / deployed \\
\midrule
Faster-WAM & 0.545 & 0.449 & 0.258 & 0.141 & 0.32 & 0.973 & 1 / 1 \\
Fast-WAM & 0.527 & 0.415 & 0.238 & 0.118 & 0.30 & 0.987 & 1 / 1 \\
UWM & 0.605 & 0.428 & 0.233 & 0.114 & 0.048 & 1.000 & 1 / 1 \\
Cosmos Policy\ph & 0.415 & 0.291 & 0.140 & 0.070 & 0.68 & 0.853 & $\frac12$ / $\frac12$ \\
LingBot-VA & 0.254 & 0.211 & 0.083 & 0.053 & 0.21 & 0.970 & 1 / 1 \\
\bottomrule
\end{tabular}
\end{table}

\paragraph{Action and timing changes.}\label{sec:s-failmodes} The shift displaces the first action chunk under every shifted condition: Recolor+Flat moves it by 0.12--0.27 per step (Table~\ref{tab:s-failmodes}a). ROOT brings it back to 0.04--0.07 of the native Original chunk while its own Original chunk differs from native by only 0.004--0.011 (Table~\ref{tab:s-failmodes}, left). Where a native model still succeeds under Recolor+Flat it is slower (Faster-WAM median 318 against 234 steps; 79\% of doubly solved layouts more than 10\% slower, Fast-WAM 93\%, UWM 57\%); under ROOT these successes take as long as in the Original scene (Faster-WAM median relative slowdown $+1.0$\%; within $\pm2$\% on Fast-WAM, Cosmos Policy and LingBot-VA) and the share slowed by more than 10\% falls to 3--7\% (UWM 23\%; Table~\ref{tab:s-failmodes}b), so the adapter changes perturbed actions and timing, not original ones.

\begin{table}[!htb]
\centering\footnotesize
\caption{\textbf{ROOT pulls shifted actions back while barely moving original ones, and restores the timing.} Left: first-action-chunk distance to the native Original chunk (median, 95\% interval); R+F: Recolor+Flat; under Recolor+Flat the first chunk already departs from its native original-scene counterpart (hollow), ROOT pulls it back (filled) while moving the original-scene chunk itself an order of magnitude less (diamonds). (a) The same distances as numbers: median per-step L2 distance over the six pose dimensions (normalized actions), 95\% bootstrap intervals. (b) Policy steps of successful episodes (median; successes in brackets) and, among layouts solved under both Original and Recolor+Flat, the share more than 10\% slower under Recolor+Flat. Horizons (policy steps) differ by model, so compare within a model only: Faster-WAM 700, Fast-WAM 700, Cosmos Policy 520, UWM 1000, LingBot-VA 796, DiT4DiT 520.}
\label{tab:s-failmodes}\label{fig:s-actions}
\begin{minipage}[t]{0.42\linewidth}\vspace{0pt}
\includegraphics[width=\linewidth]{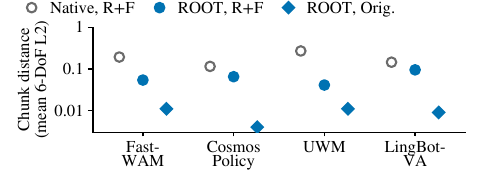}
\end{minipage}\hfill
\begin{minipage}[t]{0.56\linewidth}\vspace{0pt}\scriptsize
\emph{(a) First-chunk distance to native, Original}\par\smallskip
\begin{tabular}{@{}l ccc@{}}
\toprule
& native, R+F & ROOT, R+F & ROOT, Original \\
\midrule
Fast-WAM & 0.193 [0.173, 0.213] & 0.054 [0.049, 0.059] & 0.011 [0.010, 0.012] \\
Cosmos Policy\ph & 0.115 [0.105, 0.126] & 0.065 [0.057, 0.073] & 0.004 [0.003, 0.004] \\
UWM & 0.270 [0.201, 0.362] & 0.041 [0.036, 0.053] & 0.011 [0.008, 0.014] \\
\bottomrule
\end{tabular}
\end{minipage}

\medskip
\begin{tabular}{@{}l cc c cc c@{}}
\toprule
\multicolumn{7}{@{}l}{\emph{(b) Duration of successes (policy steps)}} \\
& \multicolumn{3}{c}{native} & \multicolumn{3}{c}{ROOT} \\
\cmidrule(lr){2-4}\cmidrule(lr){5-7}
& Original & R+F & $>$10\% slower & Original & R+F & $>$10\% slower \\
\midrule
Faster-WAM & 234 (147) & 318 (14) & 79\% [57, 100] & 233 (146) & 234.5 (144) & 4\% [1, 8] \\
Fast-WAM & 229.5 (94) & 288 (16) & 93\% [80, 100] & 232 (93) & 229 (88) & 7\% [2, 13] \\
Cosmos Policy\ph & 245 (95) & 254 (62) & 21\% [10, 31] & 245.5 (98) & 245 (99) & 3\% [0, 7] \\
UWM & 272 (111) & 311 (25) & 57\% [38, 76] & 274 (106) & 290 (96) & 23\% [15, 33] \\
\bottomrule
\end{tabular}
\end{table}

\section{Beyond SurfaceShift}
\label{sec:s-beyond}

\subsection{GE-Act on CALVIN}
\label{sec:s-ge}

CALVIN scene D has 1000 fixed chains of five tasks (length: tasks completed in a row; SR5: share of full chains); our re-run of the unmodified GE-Act gives 4.20 (reported 4.26). Chains 0--249 serve screening, 250--749 the first confirmatory batch and 750--999 the untouched batch. The conditions rewrite only table-surface pixels in the render path (segmentation of the same state, native shading kept, no physics call). The adapter is a rank-32 LoRA trained on 600 new-seed native-policy chains; the controller chose blocks 0--7 at $\alpha^\ast=1$.

\begin{table}[!htb]
\centering\footnotesize
\setlength{\tabcolsep}{3pt}
\caption{\textbf{GE-Act on CALVIN, Native$\to$ROOT.} Mean chain length and SR5 (\%). Wins: chains where ROOT completes more/fewer tasks (ties). $p$: one-sided exact sign test, Holm within each pre-registered family. NI: SR5 non-inferiority at $-10$\pp\ (Holm), lower bound of the difference; Original rows are paired through an $\alpha=0$ control arm in the same processes.}
\label{tab:s-ge}
\begin{tabular}{@{}llccccc@{}}
\toprule
Condition & Chains & $n$ & Length & SR5 & Wins & Holm $p$ \\
\midrule
Teal table & 250--749 & 500 & 3.45$\to$4.16 & 52.2$\to$73.4 & 153/34 (313) & $1.6\times10^{-19}$ \\
Marble (unseen) & 250--749 & 500 & 3.03$\to$4.17 & 37.4$\to$70.6 & 223/35 (242) & $1.1\times10^{-34}$ \\
Teal table & 750--999 & 250 & 3.35$\to$4.18 & 50.4$\to$72.0 & 78/15 (157) & $1.9\times10^{-11}$ \\
Marble (unseen) & 750--999 & 250 & 3.08$\to$4.13 & 37.2$\to$69.2 & 108/21 (121) & $4.0\times10^{-15}$ \\
\midrule
Original & 750--999 & 250 & 4.28$\to$4.32 & 76.4$\to$77.2 & \multicolumn{2}{c}{NI, lower bound $-3.1$\pp} \\
Original & 250--749 & 500 & 4.22$\to$4.18 & 73.6$\to$73.0 & \multicolumn{2}{c}{NI, lower bound $-3.1$\pp} \\
\bottomrule
\end{tabular}
\end{table}

\subsection{LIBERO-Plus}
\label{sec:s-lp}

Variants run from initial state 0, one trial per variant and arm; the clean prompt omits the benchmark's suffix (e.g.\ ``light N''), a control we added. Background: 86/11 discordant under the official prompt (Holm $p=1.4\times10^{-15}$), 90/8 under the clean prompt ($p=1.1\times10^{-18}$); Robot and Noise are non-inferior ($p=2.7\times10^{-7}$, $2.6\times10^{-11}$). ROOT beats every same-budget arm on Background (exact one-sided sign test, Holm at $0.025$; six of six) by $+11.8$ to $+18.7$\pp\ (clean) and $+17.3$ to $+21.1$\pp\ (official), while each arm falls 7--25\pp\ below native on Light, Layout, Camera, Language and Robot. The Spatial, Object and Goal gains are Holm-significant, and Cosmos Policy is superior on all 289 variants. Training-recipe methods report PICA-WAM $+25.6$\pp, $\Delta$WAM $+40.4$\pp, Robust-WAM $+10.3$\pp\ (four suites).

\begin{figure}[!htb]
\centering
\includegraphics[width=0.86\linewidth]{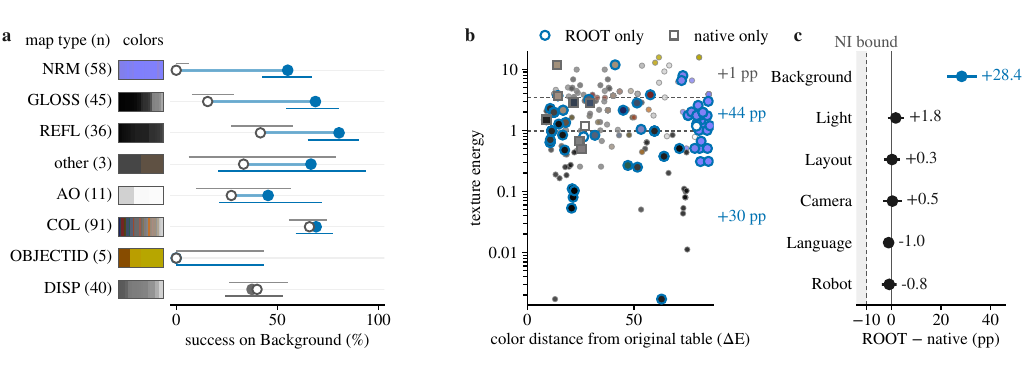}
\caption{\textbf{Where ROOT helps on LIBERO-Plus.} (a) Background success by texture-map type, official prompt. (b) The 289 Background variants by color distance and texture energy (blue: only ROOT succeeds, orange: only native). (c) Paired difference per axis, clean prompt; dashed: the $-10$\pp\ bound. Panels (a, b) are descriptive (joint model: Table~\ref{tab:s-af-lp}).}
\label{fig:s-lp}
\end{figure}

\begin{table}[!htb]
\centering\scriptsize
\setlength{\tabcolsep}{2.2pt}
\caption{\textbf{LIBERO-Plus, Faster-WAM: ROOT repairs Background and stays non-inferior elsewhere; same-budget repairs do not.} Success (\%) on LIBERO-10 subsets; Layout, Camera and Language come from a separate boundary test without chroma arm. NI: non-inferior to native at $-10$\pp\ (clean prompt); $\Delta$: ROOT$-$native, pp, score interval. Aug-BC, In-pert. and Rep.-anch. are same-budget repairs on LIBERO-Long variants with ROOT's renderings, adapter, steps and checkpoint rule; $^\ast$ROOT better (exact sign test, Holm; one-sided at 0.025, equivalent to two-sided at 0.05 with ROOT better). $^a$No suffix. $^b$Joint pose restored after the state reset. $^c$Motion, Gaussian, glass and zoom blur, fog. $^r$Re-implemented. Bold: best per row and prompt.}
\label{tab:s-lp}
\textbf{(a) Per subset, prompt and repair arm}\par\vspace{1pt}
\begin{tabular*}{\linewidth}{@{}l@{\extracolsep{\fill}}cccccc@{\hspace{5pt}}cccccc@{}}
\toprule
& \multicolumn{6}{c}{official prompt} & \multicolumn{6}{c}{clean prompt} \\
\cmidrule(lr){2-7}\cmidrule(lr){8-13}
Subset ($n$) & native & chroma & ROOT & Aug-BC & In-pert.$^r$ & Rep.-anch.$^r$ & native & ROOT & $\Delta$ & Aug-BC & In-pert.$^r$ & Rep.-anch.$^r$ \\
\midrule
Background (289) & 35.3 & 49.8 & \textbf{61.2} & 42.2$^\ast$ & 43.9$^\ast$ & 40.1$^\ast$ & 39.8 & \textbf{68.2} & --- & 49.5$^\ast$ & 56.4$^\ast$ & 49.5$^\ast$ \\
Light (274) & 78.5 & 82.8 & \textbf{88.7} & --- & --- & --- & 92.3 & \textbf{94.2}\textsuperscript{NI} & --- & 70.4$^\ast$ & 74.5$^\ast$ & 72.3$^\ast$ \\
Layout (312) & 88.1 & --- & 86.5 & --- & --- & --- & 87.5 & \textbf{87.8}\textsuperscript{NI} & $+0.3$ [$-3.0$, $+3.7$] & 76.6$^\ast$ & 80.5$^\ast$ & 78.8$^\ast$ \\
Camera (419) & 32.0 & --- & 27.5 & --- & --- & --- & 43.9 & \textbf{44.4}\textsuperscript{NI} & $+0.5$ [$-3.2$, $+4.2$] & 31.3$^\ast$ & 30.6$^\ast$ & 30.6$^\ast$ \\
Language (383)$^a$ & --- & --- & --- & --- & --- & --- & \textbf{97.7} & 96.6\textsuperscript{NI} & $-1.0$ [$-3.3$, $+1.0$] & 72.3$^\ast$ & 87.7$^\ast$ & 79.4$^\ast$ \\
Robot (393)$^b$ & --- & --- & --- & --- & --- & --- & \textbf{62.3} & 61.6\textsuperscript{NI} & $-0.8$ [$-3.8$, $+2.2$] & 52.9$^\ast$ & 55.2$^\ast$ & 50.9$^\ast$ \\
Noise (449)$^c$ & 51.4 & --- & 53.5 & --- & --- & --- & 69.5 & \textbf{72.4}\textsuperscript{NI} & $+2.9$ [$-0.4$, $+6.3$] & --- & --- & --- \\
\bottomrule
\end{tabular*}
\par\vspace{4pt}\textbf{(b) Background on other suites and Cosmos Policy (official prompt)}\par\vspace{1pt}
\begin{tabular*}{\linewidth}{@{}l@{\extracolsep{\fill}}cccc@{}}
\toprule
& Faster-WAM Spatial (258) & Faster-WAM Object (248) & Faster-WAM Goal (281) & Cosmos Policy Long (289) \\
\midrule
native & 76.4 & 74.2 & 51.6 & 69.2 \\
ROOT & \textbf{90.7} & \textbf{87.1} & \textbf{61.2} & \textbf{76.5} \\
$\Delta$, pp [95\% CI] & --- & --- & --- & $+7.3$ [2.8, 12.0], $p=0.0012$ \\
Light $\Delta$, pp (NI) & $-0.3$ & $+1.0$ & $-1.4$ & --- \\
\bottomrule
\end{tabular*}
\end{table}

\subsection{Other policies and conditions}
\label{sec:s-redobj}
\label{sec:s-gr00t}

\paragraph{Red-object.} Without object recoloring in training, ROOT improves Red-object on 150 layouts from 138 to 145 with the neutral instruction (8/1, two-sided Holm $p=0.039$) and from 136 to 145 with the conflicting one (9/0, $p=0.0078$). Cells of Table~\xr{tab:main}: blocks 0--7 solve 141 of 150 Red-object layouts (native 139, full-window adapter 144; non-inferior to native at $-10$\pp, Tango $p=1.0\times10^{-4}$); chroma correction scores 146 on Original and 141 on Red-object, late-value restoration 146 and 144 (native 147 and 139).

\paragraph{GR00T N1.7.} The released LIBERO-10 checkpoint (3.1B parameters) reads images through a frozen vision-language model into a flow-matching action DiT and was trained with color jitter; it runs on the same 150 layouts and conditions as Faster-WAM (horizon 720). Recolor+Flat is not worse than Original (132 against 128 of 150, ratio 1.03 [0.96, 1.11], pair-level bootstrap; 10 layouts lost, 14 gained, exact McNemar $p=0.54$; one kitchen task drops from 12 to 6 of 15), so there is nothing to repair and ROOT is not applied.

\section{Real-Robot Validation}
\label{sec:s-realrobot}

\paragraph{Setup and ROOT without a renderer.} \label{sec:s-rr-system}A Piper 6-DoF arm with a third-person and a wrist camera performs T1 \emph{put the yellow lemon in the bowl}, T2 \emph{put the kettle on the stove} and T3 \emph{put the white cup in the bowl and the banana to its right} (limits 30, 30, 45\,s); human contact, time-out or emergency stop fails. Native: the Faster-WAM LIBERO checkpoint fine-tuned on 120 of 150 gray-mat demonstrations (3335 steps); ten per task are held out for the checkpoint and scale rules. For ROOT, in every fourth demonstration frame the table region (color mask, fingertips excluded) is multiplied by a random color in both cameras (saturation 0.2--0.9, value 0.3--0.9) or, with probability 0.5, flattened to its mean color; hues near sky blue are excluded; no test-mat photograph is used. The rules select blocks 0--7, step 7500 and $\alpha^\ast=1$; $\alpha$ is 0 for native, 1 for ROOT.

\paragraph{Mats, placements and blinding.} The unseen mats are plain sky blue, plain black and a gray-white checkerboard (Fig.~\xr{fig:realrobot}a). Each task has 15 placements per mat; both models run each on consecutive trials in random order with the same inference seed, and the operator is blind to the model.

\begin{table}[!htb]
\centering\scriptsize
\setlength{\tabcolsep}{2pt}
\caption{\textbf{Real-robot results.} (a) Successes/trials per task and mat; color shift pools sky blue and black. (b) Success (\%) with Wilson 95\% intervals (single proportions, pairing ignored) and ROOT$-$native difference; $n$ trials per model.}
\label{tab:realrobot}
\textbf{(a) Counts}\par\vspace{1pt}
\begin{tabular*}{\linewidth}{@{}l@{\extracolsep{\fill}}cccccccccc@{}}
\toprule
& \multicolumn{2}{c}{Original} & \multicolumn{2}{c}{Sky blue} & \multicolumn{2}{c}{Black} & \multicolumn{2}{c}{Color shift} & \multicolumn{2}{c}{Checkerboard} \\
\cmidrule(lr){2-3}\cmidrule(lr){4-5}\cmidrule(lr){6-7}\cmidrule(lr){8-9}\cmidrule(lr){10-11}
Task & Native & ROOT & Native & ROOT & Native & ROOT & Native & ROOT & Native & ROOT \\
\midrule
T1 lemon & 13/15 & 14/15 & 5/15 & 14/15 & 7/15 & 15/15 & 12/30 & 29/30 & 0/15 & 2/15 \\
T2 kettle & 14/15 & 14/15 & 6/15 & 14/15 & 5/15 & 14/15 & 11/30 & 28/30 & 0/15 & 3/15 \\
T3 cup, banana & 15/15 & 15/15 & 5/15 & 14/15 & 5/15 & 13/15 & 10/30 & 27/30 & 0/15 & 1/15 \\
\midrule
All & 42/45 & 43/45 & 16/45 & 42/45 & 17/45 & 42/45 & 33/90 & 84/90 & 0/45 & 6/45 \\
\bottomrule
\end{tabular*}
\par\vspace{4pt}\textbf{(b) Percentages: all tasks, and color shift per task}\par\vspace{1pt}
\begin{tabular*}{\linewidth}{@{}l@{\extracolsep{\fill}}ccccc@{\hspace{8pt}}ccc@{}}
\toprule
& \multicolumn{5}{c}{all tasks} & \multicolumn{3}{c}{color shift, per task (30)} \\
\cmidrule(lr){2-6}\cmidrule(lr){7-9}
& Original (45) & Sky blue (45) & Black (45) & Checkerboard (45) & Color shift (90) & T1 & T2 & T3 \\
\midrule
Native & 93.3 [82.1, 97.7] & 35.6 [23.2, 50.2] & 37.8 [25.1, 52.4] & 0.0 [0.0, 7.9] & 36.7 [27.4, 47.0] & 40.0 [24.6, 57.7] & 36.7 [21.9, 54.5] & 33.3 [19.2, 51.2] \\
ROOT & \textbf{95.6} [85.2, 98.8] & \textbf{93.3} [82.1, 97.7] & \textbf{93.3} [82.1, 97.7] & \textbf{13.3} [6.3, 26.2] & \textbf{93.3} [86.2, 96.9] & \textbf{96.7} [83.3, 99.4] & \textbf{93.3} [78.7, 98.2] & \textbf{90.0} [74.4, 96.5] \\
$\Delta$ (pp) & $+2.2$ & $+57.8$ & $+55.6$ & $+13.3$ & $+56.7$ & $+56.7$ & $+56.7$ & $+56.7$ \\
\bottomrule
\end{tabular*}
\end{table}

\paragraph{Tests.} Sky blue and black: one-sided exact sign tests on the discordant placements, pooled over tasks and stratified by task, Holm over the two. Original mat: non-inferiority at $-10$\pp\ (Tango score interval, 45 pairs; lower bound above $-7.2$\pp\ for any pairing of the observed counts). Checkerboard: all six discordant placements favor ROOT (two-sided exact sign test).

\begin{figure}[!htb]
\centering
\setlength{\tabcolsep}{0pt}
\begin{tabular}{@{}c@{}}
{\scriptsize Sky blue, T1, trial T081. Native: failure (missed grasp), 6.3\,s}\\[-1pt]
\includegraphics[width=\linewidth]{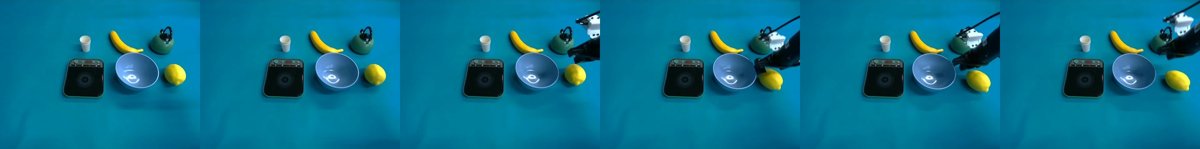}\\[0pt]
\includegraphics[width=\linewidth]{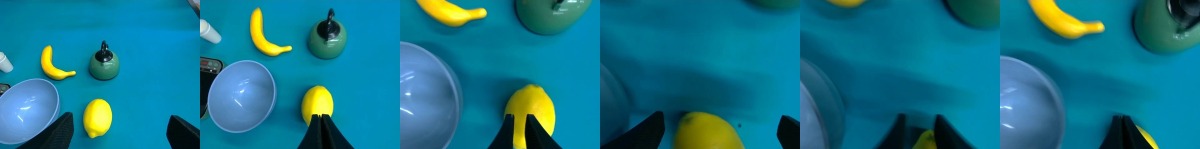}\\[4pt]
{\scriptsize Sky blue, T1, trial T077. ROOT: success, 12.4\,s}\\[-1pt]
\includegraphics[width=\linewidth]{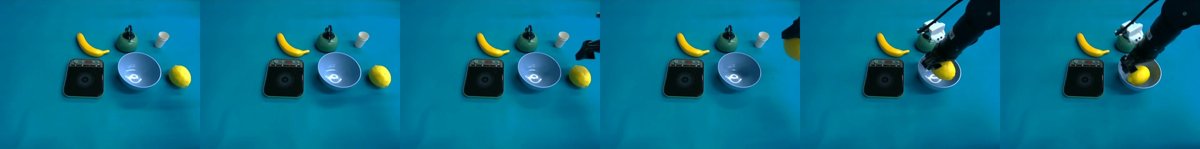}\\[0pt]
\includegraphics[width=\linewidth]{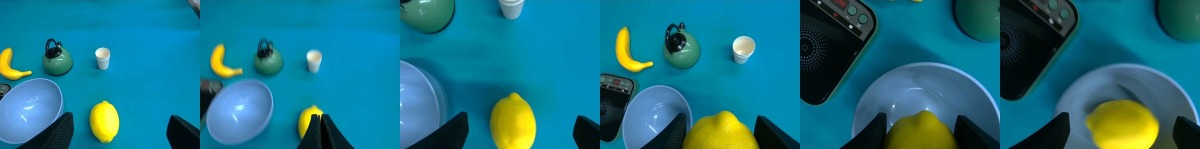}\\[4pt]
{\scriptsize Black, T2, trial T287. Native: failure (missed grasp), 5.1\,s}\\[-1pt]
\includegraphics[width=\linewidth]{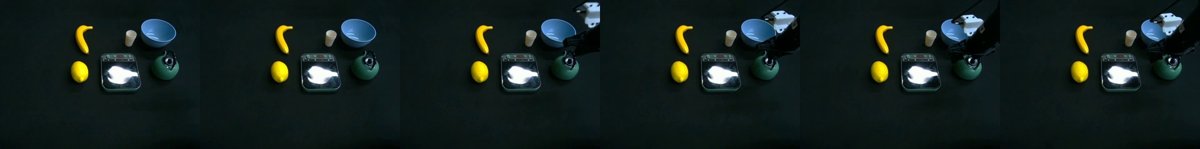}\\[0pt]
\includegraphics[width=\linewidth]{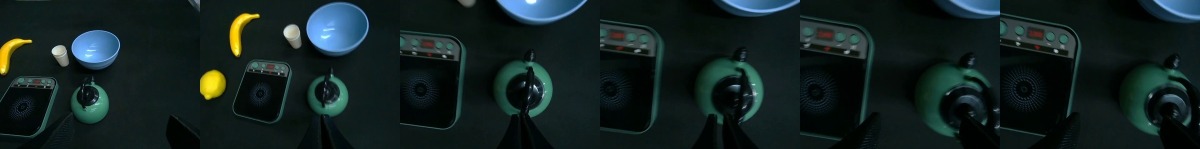}\\[4pt]
{\scriptsize Black, T2, trial T298. ROOT: success, 12.0\,s}\\[-1pt]
\includegraphics[width=\linewidth]{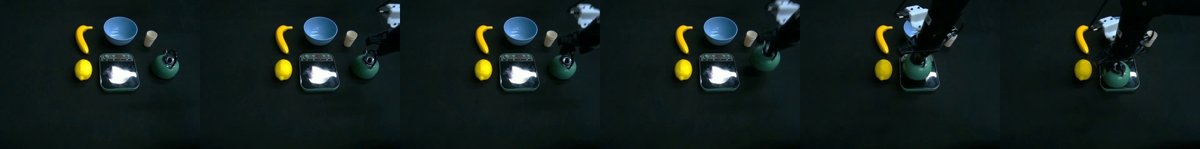}\\[0pt]
\includegraphics[width=\linewidth]{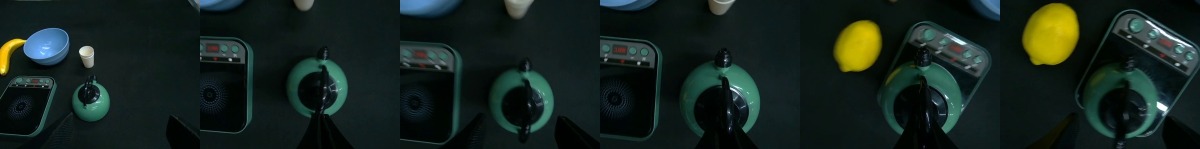}
\end{tabular}
\caption{\textbf{Fixed third-person camera (top) and wrist camera (bottom) on the sky-blue and black mats.} Six frames per trial at the recorded times of the video.}
\label{fig:s-rr-roll}
\end{figure}

\begin{figure}[!htb]
\centering
\setlength{\tabcolsep}{0pt}
\begin{tabular}{@{}c@{}}
{\scriptsize Checkerboard, T3. Native: failure (missed grasp), 9.6\,s}\\[-1pt]
\includegraphics[width=\linewidth]{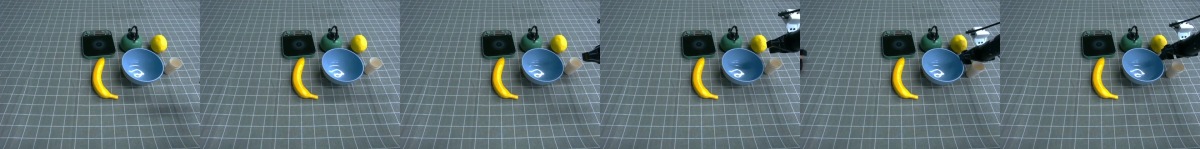}\\[0pt]
\includegraphics[width=\linewidth]{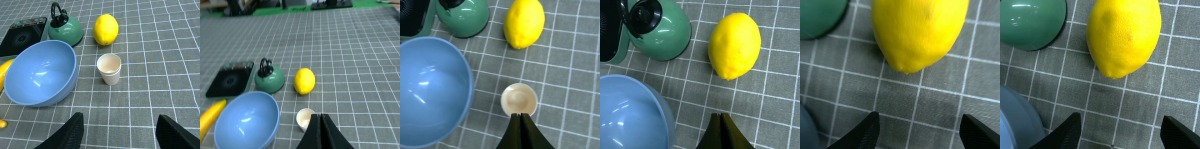}\\[4pt]
{\scriptsize Checkerboard, T3. ROOT: success, 22.8\,s}\\[-1pt]
\includegraphics[width=\linewidth]{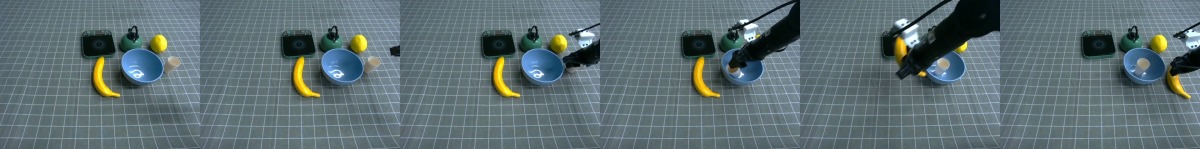}\\[0pt]
\includegraphics[width=\linewidth]{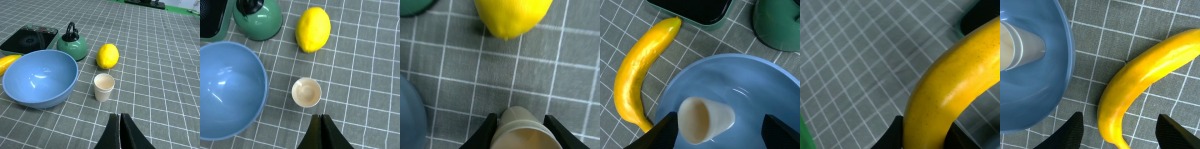}
\end{tabular}
\caption{\textbf{Fixed third-person camera (top) and wrist camera (bottom) on the checkerboard mat.} Six frames per trial at the recorded times of the video.}
\label{fig:s-rr-roll-p}
\end{figure}

\section{Figure Data, Compute and Broader Impact}
\label{sec:s-fignotes}

\begin{itemize}
\item \textbf{Fig.~\xr{fig:teaser}.} Recolor+Flat is teal in the four kitchen tasks and near-black in the six others (S1.1); inset $n$: 150 layouts (Faster-WAM, UWM), 100 (Fast-WAM, Cosmos Policy). Data: Table~\ref{tab:s-main-full}.
\item \textbf{Fig.~\xr{fig:method}.} Faster-WAM; blue: blocks 0--7 of 29, orange: late video values, blocks 16--28.
\item \textbf{Fig.~\xr{fig:factors}.} (a) within-task effects; (b) 563 LIBERO-Plus Background and Light variants (Table~\ref{tab:s-af-c}).
\item \textbf{Fig.~\xr{fig:readout}.} Orange: share of the Recolor+Flat deviation removed by restoring the listed blocks from the original-scene render (offline, privileged; S3.3); blue and diamond: Table~\ref{tab:s-windows}.
\item \textbf{Fig.~\xr{fig:scale}.} DiT4DiT omitted (native kept at the window stage); data: Table~\ref{tab:s-dose}.
\item \textbf{Fig.~\xr{fig:boundary}.} (a) Chains 750--999; (b) 289 (Long), 258, 248, 281 Background variants; (c) clean prompt (Tables~\ref{tab:s-ge}, \ref{tab:s-lp}).
\item \textbf{Fig.~\xr{fig:realrobot}.} Counts: Table~\ref{tab:realrobot}.
\end{itemize}

Table~\ref{tab:s-cost} lists adapter sizes, training time and the overhead of our unmerged implementation, measured within a
run on Original and Red-object. The window probes cost 8.2 (Faster-WAM, 16 probes), 7.2 (Fast-WAM, 16) and 6.3 GPU-hours
(Cosmos Policy, 6), and the scale rule about 1 GPU-hour per model. The update folds into the base weights
($W'=W+\alpha\,(\kappa/r)BA$) for Faster-WAM, Fast-WAM and UWM; the Cosmos Policy adapter acts only on some tokens and LingBot-VA's only
in video-mode calls, so neither folds. In a benchmark of 100 inputs per model, folded latency was 0.999--1.000 of
native on the three foldable models against 1.011--1.034 unfolded. Adapters have three training seeds (all deployed models; narrow windows on Faster- and Fast-WAM; the prospective runs); Faster-WAM, Fast-WAM and Cosmos
Policy also have two inference seeds.

\begin{table}[!htb]
\centering\scriptsize
\caption{\textbf{ROOT adds 0.16--1.7\% parameters and at most 10.3\% unmerged overhead.} LoRA parameters and share of the
policy network; training wall clock on one A100 (run wall clock including data loading and held-out evaluation in
brackets); overhead of the unmerged adapter vs.\ native in the same run, per-query wall clock unless marked $^m$ (model
latency). Windows marked $w$ are the input-side windows selected by the procedure.}
\label{tab:s-cost}
\begin{tabular}{@{}l r r l l c@{}}
\toprule
Model / window & LoRA params & \% policy & training & overhead & foldable \\
\midrule
Faster-WAM, blocks 0--28 & 55.1M & 0.99 & 5.8\,h (7.8\,h), 40\,GB & $+$2.7 to $+$4.3\% & yes \\
Faster-WAM, blocks 0--7$^w$ & 15.2M & 0.27 & 5.5\,h (6.7\,h), 40\,GB & --- & yes \\
Fast-WAM, blocks 0--29 & 57.0M & 0.95 & 3.4\,h (5.3\,h), 80\,GB & $+$3.8\%$^m$ & yes \\
Fast-WAM, blocks 0--7$^w$ & 15.2M & 0.25 & 3.6\,h (4.9\,h), 40\,GB & --- & yes \\
UWM, image encoder & 0.62M & 0.16 & 1.2\,h (2.2\,h), 40\,GB & $+$1.1\% & yes \\
Cosmos Policy, all blocks\ph & 33.0M & 1.69 & 11.4\,h (12.4\,h), 80\,GB & $+$10.3\%$^m$ ($\alpha=\frac12$) & no \\
Cosmos Policy, blocks 0--13$^w$ & 16.5M & 0.84 & 10.4\,h (11.4\,h), 40\,GB & --- & no \\
LingBot-VA, all blocks & 11.8M & 0.23 & $\le$6.6\,h run, shared GPU & $+$4 to $+$6\% & no \\
\bottomrule
\end{tabular}
\end{table}

\paragraph{Broader impact.} Restoring a policy's behavior under work-surface appearance change, and checking that the original
scene is preserved before deploying, can make robot policies more reliable on new surfaces. SurfaceShift evaluation code and
part of the training code, and then all code, will be released upon publication.

\end{document}